\PassOptionsToPackage{table,dvipsnames}{xcolor}
\documentclass[]{selfevolagent}

\usepackage{microtype}
\usepackage{amsmath,amssymb,amsfonts}
\usepackage{graphicx}
\usepackage{booktabs}
\usepackage{multirow}
\usepackage{bm}
\usepackage{array}
\usepackage{enumitem}
\usepackage{colortbl}
\usepackage{algorithm}
\usepackage{algcompatible}
\usepackage{subcaption}
\usepackage{makecell}
\usepackage{float}
\usepackage{wrapfig}
\usepackage{longtable}
\usepackage{tabularx}
\usepackage{xltabular}
\usepackage{url}
\usepackage{fontawesome5}
\microtypesetup{expansion=false}

\renewenvironment{figure*}{\begin{figure}}{\end{figure}}
\renewenvironment{table*}{\begin{table}}{\end{table}}

\makeatletter
\def\blfootnote{\xdef\@thefnmark{}\@footnotetext}
\makeatother

\title{Video-MemComp: Taylor-Guided Sensitivity-Aware Memory Management for Streaming Video Understanding}

\author[1]{Yonghui Niu}
\author[1]{Yuhao Ji}
\author[1]{Chang Nie}
\author[2]{Jiao Zhang}
\author[2]{Yilong Li}
\author[2,\dagger]{\protect\\ Junlan Feng}
\author[3]{Ran He}
\author[1]{Caifeng Shan}
\author[1,\dagger]{Chaoyou Fu}

\affiliation[1]{NJU}
\affiliation[2]{Jiutian AI}
\affiliation[3]{CASIA}

\abstract{
The rapid growth of video data has created an urgent need for efficient video understanding systems that can handle long and continuous visual inputs, posing significant challenges for Multimodal Large Language Models (MLLMs) due to severe memory bottlenecks. Moreover, existing compression methods often overlook the explosive accumulation of approximation errors. To address this issue, we propose \textbf{Video-MemComp}, a training-free and sensitivity-aware context management framework for memory-constrained streaming video understanding. Our method theoretically decomposes compression error into orthogonal \textit{local approximation error} and \textit{layer-wise structural sensitivity} via a first-order Taylor expansion. To minimize \textit{local approximation error}, we perform token aggregation in a pre-RoPE safe zone with an attention-conserving logit bias, mitigating geometric distortions caused by RoPE rotations. To manage \textbf{layer-wise structural sensitivity}, we dynamically allocate the KV cache budget based on offline gradient priors and online context entropy, prioritizing high-sensitivity layers. Video-MemComp outperforms or achieves performance comparable to full-cache and state-of-the-art compression methods on benchmarks including MLVU, VideoMME, and EgoSchema. Notably, it improves the overall accuracy of Qwen2.5-VL-3B from 62.3\% to 64.1\% while reducing the KV cache to an equivalent of 6,000 tokens per layer, saving over 90\% of KV memory.
}

\metadata[\faEnvelope\ Corresponding authors]{\email{fengjunlan@chinamobile.com},\ \email{chaoyoufu@nju.edu.cn}}
\metadata[Project Page]{\url{https://hhnk666.github.io/Video-MemComp.github.io/}}
\metadata[GitHub]{\url{https://github.com/hhnk666/videomemorycompress}}

\begin{document}
\maketitle
\blfootnote{This paper has been accepted by ACM Multimedia (MM) 2026.}

\section{Introduction}
\label{sec:intro}
\begin{figure}[t]
    \centering
    \includegraphics[width=\linewidth]{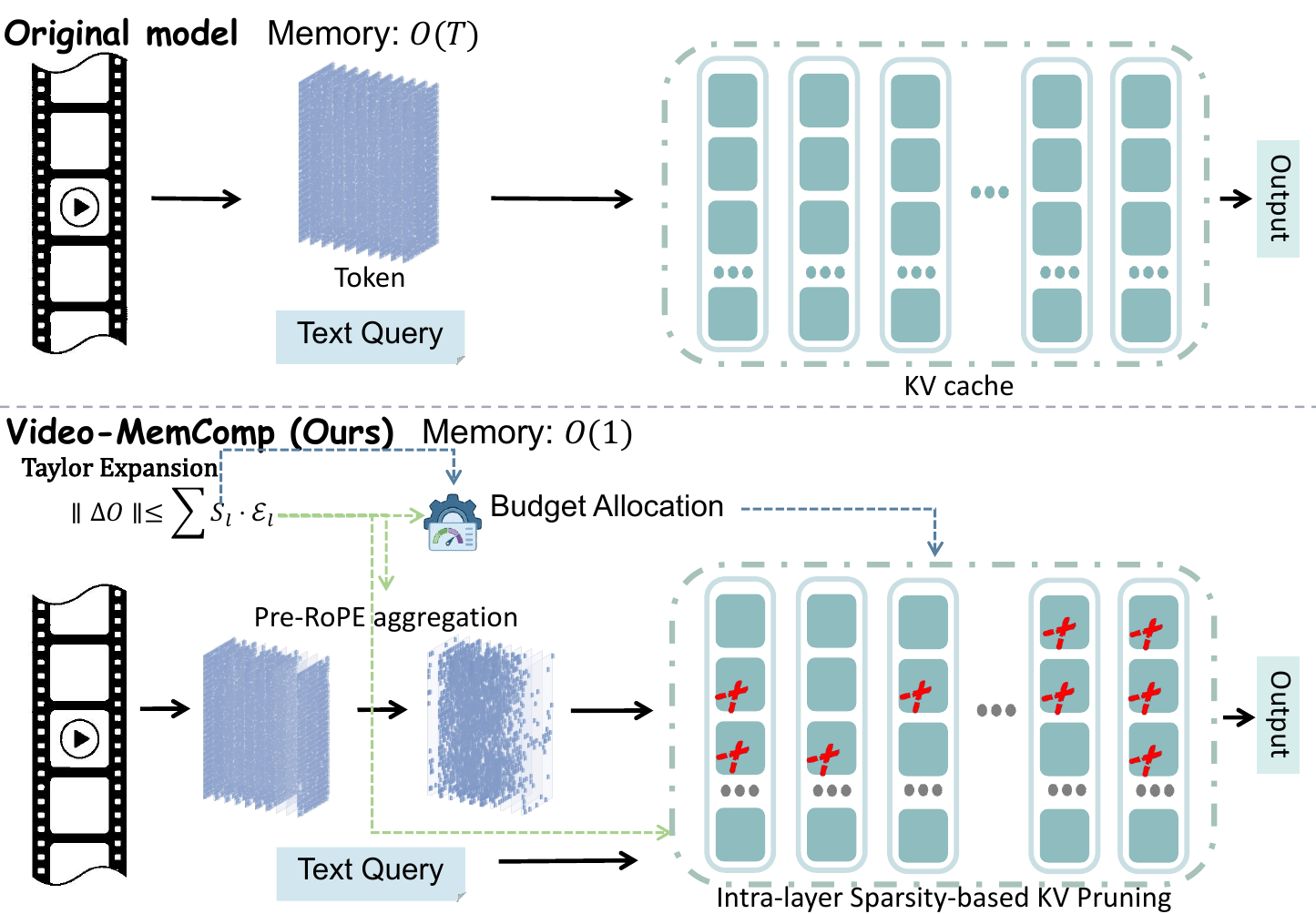}
\caption{\textbf{Our architecture}. While standard Video-LLM models suffer from linear $O(T)$ memory growth, Video-MemComp maintains strictly $O(1)$ memory via a Taylor-expansion-guided hybrid strategy, enabling efficient streaming video understanding. Specifically, we decompose the error via Taylor expansion into local approximation error $\mathcal{E}_l$ and layer-wise structural sensitivity $\mathcal{S}_l$. To minimize $\mathcal{E}_l$, we adopt Attention-Preserved Pre-RoPE Aggregation and Intra-layer Sparsity-based KV pruning. Meanwhile, $\mathcal{S}_l$ is used to guide layer-wise budget allocation.}

    \label{fig:methodoverall}
\end{figure}

As Multimodal Large Language Models (MLLMs) evolve from static image understanding to long-form streaming video reasoning, they exhibit strong emergent capabilities in temporal reasoning and narrative comprehension~\cite{openai20234v, liu2023visual, bai2025qwen2}. However, the paradigm is quickly shifting towards real-time, continuous multimedia applications, such as AR/VR smart glasses, embodied agents, and intelligent surveillance systems. In these streaming video scenarios, video assistants must continuously perceive infinite visual streams while enabling seamless, low-latency human-computer interaction. This imposes a severe conflict: the streaming multimedia input is temporally unbounded, yet the system's computational resources and GPU memory are strictly finite. A fatal system bottleneck emerges as the Key-Value (KV) cache grows linearly with the sequence length ($O(T)$), making continuous real-time inference highly susceptible to Out-Of-Memory failures, especially on resource-constrained edge devices or consumer-grade GPUs~\cite{shao2025tokens, ning2025livevlm, chen2025streamingtom}.

To alleviate memory pressure, recent works have explored memory compression techniques, including heuristic pruning and token merging. However, these approaches exhibit limitations in addressing the error accumulation mechanism, where minute approximation errors in specific layers (typically shallow ones) can amplify in the final output. Specifically, pruning methods, whether utilizing static budgets~\cite{zhang2023h2o, li2024snapkv} or adaptive allocation based on information density~\cite{shen2025lava, feng2024ada}, often fail to fully account for the non-uniform error distribution across layers. Conversely, merging strategies~\cite{bolya2022token, shao2025holitom} risk attention score dilution through simple averaging. Furthermore, direct KV-space merging~\cite{yang2025streammem} may overlook the properties of RoPE, where positional rotations distort the similarity metrics required for effective merging.

To address these limitations, we employ a first-order Taylor expansion to analytically approximate the disparity in how different layers impact the final model performance.
Letting $\mathcal{O}$ denote the final model output and $A^{(l)}$ represent the attention output of the $l$-th layer, this approach enables us to decompose the global output error $\|\Delta \mathcal{O}\|$ down to the layer level, revealing two orthogonal components: \textit{Layer-wise Structural Sensitivity} $\mathcal{S}_l$ and \textit{Local Approximation Error} $\mathcal{E}_l$:
\begin{equation}
    \label{eq:intro_taylor}
    \|\Delta \mathcal{O}\| \le \sum_{l=0}^{L-1} \underbrace{\|\frac{\partial \mathcal{O}}{\partial A^{(l)}}\|}_{\mathcal{S}_l} \cdot \underbrace{\|\Delta A^{(l)}\|}_{\mathcal{E}_l},
\end{equation}
where $\mathcal{E}_l$ measures the information loss from compressing layer $l$, and $\mathcal{S}_l$ quantifies the gradient-based amplification factor of that layer.
This analytical framework dictates our core design principle: we must minimize the local approximation error $\mathcal{E}_l$ under the premise of explicitly managing the layer-wise structural sensitivity $\mathcal{S}_l$.

In this work, we introduce \textbf{Video-MemComp}, a training-free and model-agnostic structure-aware hybrid framework guided by this analytical framework and designed for extreme $O(1)$ memory efficiency (see Fig.~\ref{fig:methodoverall}).
Our system design strictly maps to the two error components identified in Eq.~(\ref{eq:intro_taylor}). 
To minimize the local approximation error $\mathcal{E}_l$, we identify that the RoPE-induced geometric barrier is the primary obstacle to effective merging; consequently, we shift the aggregation process to the Pre-RoPE ``safe zone,'' utilizing an attention-conserving logit bias to strictly preserve attention mass during physical token consolidation. 
Simultaneously, to manage the structural sensitivity $\mathcal{S}_l$, we recognize that simply knowing $\mathcal{S}_l$ is insufficient and formulate memory management as a dynamic resource scheduling problem to maintain a constant-time memory footprint. We introduce a layer-wise budget allocation mechanism that leverages offline gradient priors to solve this optimization, aggressively reclaiming memory from low-sensitivity layers to favor high-sensitivity ones.

\begin{figure}[t]
    \centering
    \includegraphics[width=0.7\linewidth]{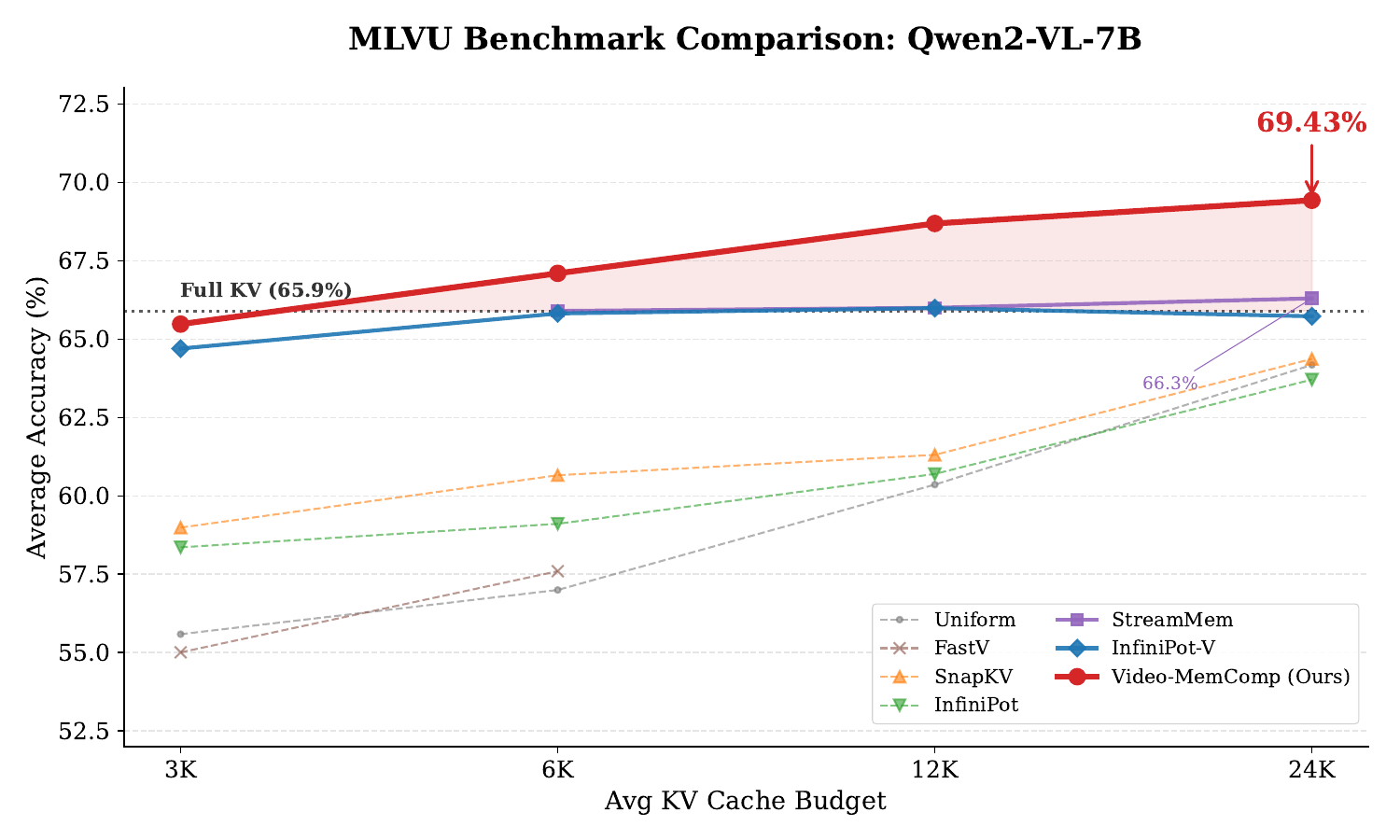}
    \caption{Performance on MLVU with Qwen2-VL-7B. Our method outperforms strong baselines across all memory budgets, and even surpasses the Full KV baseline at 6K, 12K, and 24K KV budgets, showing that Video-MemComp removes redundancy while preserving key semantics. See Section E.1 in the supplementary material for details.
}

    \label{fig:mlvu}
\end{figure}

Extensive experiments on benchmarks such as MLVU~\cite{zhou2025mlvu}, Video-MME~\cite{fu2025video}, and EgoSchema~\cite{mangalam2023egoschema} demonstrate a ``less is more'' effect: by mathematically filtering redundancy, our method matches or surpasses full-cache baselines (see Fig.~\ref{fig:mlvu}).
Notably, with Qwen2.5-VL-3B~\cite{bai2025qwen2}, Video-MemComp improves the overall accuracy from 62.3\% to 64.1\%, while limiting the average KV cache to 6,000 tokens per layer, cutting KV cache memory by over 90\%.


Our main contributions are summarized as follows:
\begin{itemize}
    \item We analyze the error accumulation mechanism in MLLMs and provide an analytical decomposition of the streaming memory compression objective into layer-wise structural sensitivity $\mathcal{S}_l$ and local approximation error $\mathcal{E}_l$ via a first-order Taylor expansion.
    \item We propose Video-MemComp, a hybrid framework that minimizes local error via Pre-RoPE aggregation and mitigates error amplification through a sensitivity-aware budget allocation strategy.
    \item We demonstrate that our Taylor-guided $O(1)$-memory strategy performs competitively with or surpasses both full-cache baselines and state-of-the-art compression methods across multiple MLLM architectures and benchmarks.
\end{itemize}

    \begin{figure*}[t]
    \centering
    \includegraphics[width=1.0\textwidth]{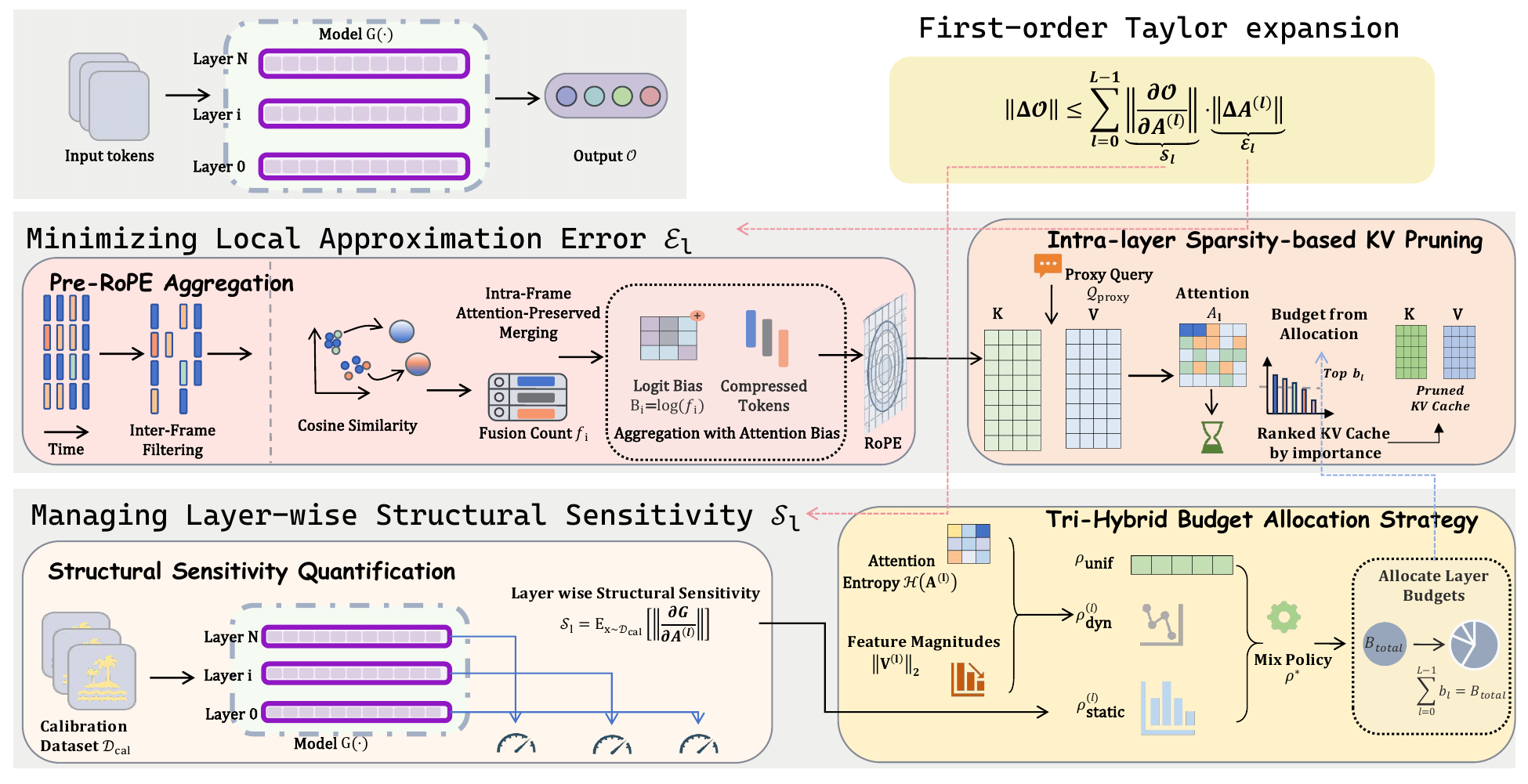} 

    \caption{Overview of our Video-MemComp Framework.
Guided by Taylor expansion (top), we split global memory compression into two objectives. 
1) Minimizing Local Approximation Error $\mathcal{E}_l$ (middle): \textit{Pre-RoPE Aggregation} merges redundant tokens based on inter-frame filtering and intra-frame merging. \textit{Intra-layer KV Pruning} selects key-value pairs using a dialogue-aware Proxy Query $\mathcal{Q}_{\text{proxy}}$. 
2) Managing Structural Sensitivity $\mathcal{S}_l$ (bottom): A \textit{Tri-Hybrid Budget Allocation} dynamically assigns the total memory budget $B_{\text{total}}$ using offline structural sensitivity, online information density, and uniform regularization to protect critical layers from aggressive pruning.}

    \label{fig:framework}
\end{figure*}

\section{Related Work}

High-resolution, long-form MLLMs \cite{zhang2024video,wang2025internvl3,li2024llava,bai2025qwen3,wang2024mpo} face memory bottlenecks from accumulating tokens and KV caches \cite{chen2024videollm,yao2025timechat}, driving solutions in token compression and KV optimization.

\noindent \textbf{Visual Token Compression in MLLMs.}
Spatial methods reduce intra-frame redundancy~\cite{shang2025llava, alvar2025divprune, zhang2025simignore, lei2025core,zhang2026d2pruner,he2026energy} but often overlook temporal redundancy~\cite{shao2025tokens}. Spatio-temporal approaches exploit inter-frame similarity or token merging~\cite{shen2024longvu, wang2024retake, fu2024framefusion, cha2024honeybee}, yet mostly rely on heuristics. Some partially violate attention conservation~\cite{liu2025video, chen2025streamingtom}, and logit-bias compensation~\cite{zeng2025token, zhao2025tinychemvl} is typically static-centric and does not resolve $O(T)$ growth, limiting guarantees on local approximation error in streaming QA.

\noindent \textbf{KV Cache Optimization for Long-Context Inference.}
The linearly growing KV cache poses a major bottleneck for long-video understanding. Existing approaches in LLMs primarily rely on heuristic strategies, such as fixed sparsity patterns, stabilization mechanisms, cumulative attention, or importance-based metrics~\cite{han2024lm, xiao2023efficient, zhang2023h2o, li2024snapkv, devoto2024simple, park2025keydiff, su2025calidrop}. 
Meanwhile, non-uniform budgeting methods dynamically adjust layer-wise allocation based on signals such as attention entropy or feature norms~\cite{feng2024ada,qin2025cakecascadingadaptivekv, shen2025lava,cai2025pyramidkvdynamickvcache,wang2026crystal}. 
More recently, HERMES~\cite{zhang2026hermes}, built upon a mechanistic investigation of attention, conceptualizes the KV cache as a hierarchical memory framework that captures video information across multiple granularities. However, existing methods often overlook layer-wise structural sensitivity, where approximation errors can accumulate and amplify in specific layers~\cite{li2025mminference, tu2024vl, wang2025sparsemm, liu2025mixing, zhang2026hermes}.

\noindent \textbf{Streaming-Oriented Memory Management.}
Retrieval and store-then-retrieve schemes incur linear storage~\cite{zhang2024flash, qian2025dispider, di2025rekv, chen2025streamingtom, patel2025cacheflow}, while bounded methods sometimes ignore inter-frame similarity or rely on heuristics~\cite{yang2025streammem, qin2025sharp}. Even $O(1)$ approaches can suffer from information loss~\cite{ning2025livevlm}, limiting their ability to allocate memory by structural sensitivity or minimize local approximation error.

Unlike these approaches, Video-MemComp utilizes Taylor expansion to theoretically decompose the compression objective into local approximation error and layer-wise structural sensitivity, ensuring a strict $O(1)$ memory footprint.

\section{Methodology}
\subsection{Problem Reformulation via Taylor Expansion}
\label{sec:method_theory}

Under the constraint of maintaining $O(1)$-memory overhead, directly optimizing the global output error $\|\Delta \mathcal{O}\|$ is intractable due to the high-dimensional non-linearity of the model. To derive a feasible surrogate, we shift our analytical focus to the attention layer outputs $A^{(l)}$, which serve as the direct interface for KV cache compression. Here, $L$ represents the total number of layers in the model and $l \in \{0, \dots, L-1\}$ denotes the layer index. 

We define $\mathcal{A}=\{A^{(0)}, \dots, A^{(L-1)}\}$ as the collection of the states. By abstracting the model output as a multivariate function $\mathcal{O}=G(\mathcal{A})$, the error $\|\Delta \mathcal{O}\|$ can be characterized layer-wise. When compression is applied, it introduces a local perturbation $\Delta A^{(l)}$ at the $l$-th layer. Assuming the perturbations are sufficiently small, we can linearize the global output deviation $\|\Delta \mathcal{O}\|$ using a first-order multivariate Taylor expansion (see Appendix \ref{app:theory_details} for details):
\begin{equation}
\label{eq:taylor_proof}
\begin{aligned}
    \|\Delta \mathcal{O}\| &= \| G(\mathcal{A} + \Delta \mathcal{A}) - G(\mathcal{A}) \| \\
    &\approx \left\| \sum_{l=0}^{L-1} \frac{\partial G}{\partial A^{(l)}} \cdot \Delta A^{(l)} \right\| \\
    &\le \sum_{l=0}^{L-1} \left\| \frac{\partial G}{\partial A^{(l)}} \right\| \cdot \left\| \Delta A^{(l)} \right\| = \sum_{l=0}^{L-1} \mathcal{S}_l \cdot \mathcal{E}_l.
\end{aligned}
\end{equation}
Eq.~(\ref{eq:taylor_proof}) provides the analytical motivation for our \mbox{Video-MemComp} framework (see Fig.~\ref{fig:framework}), guiding its hierarchical design strategy.
The local approximation error $\mathcal{E}_l = \|\Delta A^{(l)}\|$ indicates that compression must be optimized \textit{locally} within a specific layer to preserve semantic fidelity. To retain high information-density features under limited budgets, we integrate pre-RoPE aggregation with intra-layer sparsity-based KV cache pruning, effectively minimizing $\mathcal{E}_l$. Meanwhile, the layer-wise structural sensitivity $\mathcal{S}_l = \|\frac{\partial G}{\partial A^{(l)}}\|$ acts as a global amplification factor, making budget allocation a \textit{layer-wise} resource scheduling problem. Layers with high $\mathcal{S}_l$ (e.g., shallow layers) propagate errors more strongly and therefore require a larger share of the total memory budget to suppress the global error bound.

\subsection{Minimizing Local Approximation Error \texorpdfstring{$\mathcal{E}_l$}{E_l}}
\label{sec:local_error}
\noindent \textbf{Pre-RoPE Aggregation.} 
To reduce $\mathcal{E}_l$, we build upon prior KV cache compression methods \cite{yang2025streammem} but fundamentally rethink redundancy handling. Since the injection of Rotary Positional Embeddings (RoPE) entangles positional information with semantic similarity during the prefill stage, merging within the KV cache becomes geometrically unreliable. Therefore, we shift similarity-based aggregation to the Pre-RoPE stage. Furthermore, we compensate for the attention score shift caused by token aggregation, thereby minimizing $\mathcal{E}_l$ as much as possible.

Based on the principle of attention conservation, we merge similar tokens and subsequently correct the attention scores to ensure faithful token aggregation (see Appendix \ref{app:case1} for details). Our goal is to minimize information loss during compression. 
Considering that the Transformer’s core self-attention mechanism fundamentally measures relevance via the dot product between Query and Key vectors, which induces a directional matching bias, we adopt cosine similarity as the metric for redundancy. Intuitively, if two tokens share the same direction in feature space, their response patterns to any Query will be highly similar. We provide a mathematical justification for this choice in Appendix \ref{app:metric_optimality}.

To eliminate similarity redundancy at the input, we design a three-stage process (Algorithm \ref{alg:token_compression}). We first leverage temporal inertia to filter out inter-frame static redundancy. Subsequently, for intra-frame spatial redundancy, we employ an iterative bipartite matching strategy. 
Most critically, to prevent the Attention Dilution caused by heuristic merging, we track the effective fusion count $f_i$ for each surviving token (representing the number of raw tokens it encapsulates) and inject a logit bias correction $\mathbf{B}_{i} = \log(f_i)$ into the attention logits. This strictly enforces the conservation of attention mass (See Appendix \ref{app:case1} for details), ensuring that aggregated background tokens retain their collective semantic weight, thereby minimizing the local error $\mathcal{E}_l$. 

\noindent \textbf{Intra-layer Sparsity-based KV cache Pruning.} Selective pruning is applied to the KV cache in specific layers to mitigate sparsity-based redundancy. To estimate the importance of cached keys without future knowledge, we construct a Proxy Query $\mathcal{Q}_{\text{proxy}}$ using a generic dialogue template \cite{yang2025streammem}. The importance score $\mathbf{s}^{(l)}$ for layer $l$ is computed as:
\begin{equation}
    \mathbf{s}^{(l)} = \text{MeanPool}\left( \text{Softmax}\left( \frac{\mathcal{Q}_{\text{proxy}} (\mathbf{K}^{(l)})^\top}{\sqrt{d}} \right) \right).
\end{equation}
We retain only the top-$b^{(l)}$ KV pairs with the highest importance scores, where the layer-specific budget $b^{(l)}$ is dynamically determined by our inter-layer allocation strategy.

\begin{algorithm}[t]
\caption{Attention-Preserved Token Aggregation}
\label{alg:token_compression}
\begin{algorithmic}[1]
    \REQUIRE Current frame $\mathbf{X}_t$, Previous frame $\mathbf{X}_{t-1}$
    \REQUIRE Thresholds: $\theta_{\text{inter}}$, $\theta_{\text{intra}}$; Gate $\kappa=0.5$
    \REQUIRE Temperature $\tau=1/150$
    \ENSURE Compressed tokens $\mathbf{Y}$, Fusion counts $\mathbf{f}$
    
    \STATE \textit{Detailed implementation available in Appendix \ref{app:implementation_details}} 
    
    \STATE \textbf{Phase 1: Inter-Frame Filtering}
    \STATE $\mathbf{s}_{\text{time}} \leftarrow \text{CosSim}(\mathbf{X}_t, \mathbf{X}_{t-1})$
    \STATE $\mathcal{I}_{\text{raw}} \leftarrow \{ i \mid \mathbf{s}_{\text{time}}[i] < \theta_{\text{inter}} \}$ \hfill \textcolor{gray}{// Keep dynamic tokens}
    \STATE $\mathbf{X}_{\text{raw}} \leftarrow \mathbf{X}_t[\mathcal{I}_{\text{raw}}]$
    
    \STATE \textbf{Phase 2: Bipartite Matching Decision}
    \STATE Initialize survivor set $\mathbf{X}_{\text{surv}} \leftarrow \mathbf{X}_{\text{raw}}$
    \FOR{iter $k = 1$ \textbf{to} $K$}
        \STATE Partition $\mathbf{X}_{\text{surv}}$ into sets $\mathcal{A}$ (source) and $\mathcal{B}$ (target)
        \STATE Compute edges: $(i, j)$ maximizing $\text{CosSim}(a_i, b_j) > \theta_{\text{intra}}$
        \STATE Update survivors: $\mathbf{X}_{\text{surv}} \leftarrow \mathcal{B} \cup \{ a \in \mathcal{A} \mid a \text{ is unmatched} \}$
    \ENDFOR
    
    \STATE \textbf{Phase 3: Computing Fusion Counts }

    \STATE Let $M = |\mathbf{X}_{\text{surv}}|$, $N = |\mathbf{X}_{\text{raw}}|$
    
    \STATE \textcolor{gray}{// 3.1 Gated Similarity Matrix}
    \STATE $\mathbf{S} \leftarrow \mathbf{X}_{\text{surv}} \cdot (\mathbf{X}_{\text{raw}})^\top$
    \STATE $\mathbf{G}_{ij} \leftarrow \mathbf{S}_{ij}$ if $\mathbf{S}_{ij} \ge \kappa$ else $-\infty$
    
    \STATE \textcolor{gray}{// 3.2 Attention Weights \& Fusion Counts}
    \STATE $\mathbf{M} \leftarrow \text{Softmax}(\mathbf{G} / \tau, \text{dim}=0)$
    \STATE $\mathbf{f} \leftarrow \sum_{j=1}^N \mathbf{M}_{:, j}$ \hfill \textcolor{gray}{// Sum over columns (raw tokens)}
    
    \STATE \textcolor{gray}{// 3.3 Feature Update}
    \STATE $\mathbf{W}_{ij} \leftarrow \mathbf{M}_{ij} / ( \sum_{k=1}^N \mathbf{M}_{ik} + \epsilon)$
    \STATE Reconstruct: $\mathbf{Y} \leftarrow \mathbf{W} \cdot \mathbf{X}_{\text{raw}}$
    
    \STATE \textbf{return} $\mathbf{Y}, \mathbf{f}$
\end{algorithmic}
\end{algorithm}

\subsection{Minimizing Local Approximation Error \texorpdfstring{$\mathcal{E}_l$}{E_l}}

\label{sec:budget_allocation}

In Eq.~\ref{eq:taylor_proof}, we established that the upper bound of the total error is determined by both structural sensitivity and local approximation error. To minimize performance loss under a total budget constraint $B_{\text{total}}$, we formalize layer-wise budget allocation as a constrained optimization problem.

\noindent \textbf{Structural Sensitivity Quantification.} 
Structural sensitivity $\mathcal{S}_l = \| \partial G / \partial A^{(l)} \|$ quantifies the topological importance of layer $l$. Layers with larger gradient norms suggest that minor local deviations $\Delta A^{(l)}$ will be cascaded and amplified by subsequent layers, which significantly affects the accuracy of the final prediction.

We quantify this property through offline calibration. 
Specifically, using a representative dataset $\mathcal{D}_{\text{cal}}$ (e.g., LLaVA-Pretrain~\cite{liu2023visual}), 
we estimate the expected gradient norm of each token by performing backpropagation over the dataset:
\begin{equation}
\label{eq:cal}
    \mathcal{S}_l = \mathbb{E}_{\mathbf{x} \sim \mathcal{D}_{\text{cal}}} \left[ \left\| \frac{\partial G}{\partial A^{(l)}} \right\| \right].
\end{equation}
This offline-computed $\mathcal{S}_l$ serves as a robust {Static Prior}, guiding the budget toward structurally critical layers.

\noindent \textbf{Optimization Objective.}
Let $b_l$ denote the KV cache budget allocated to layer $l$, and $\mathcal{E}^{(l)}(b_l)$ be the local approximation error (measured in Frobenius norm) induced by pruning under this budget. Our goal is to find an optimal budget distribution $\mathbf{B}^* = \{B_0, \dots, B_{L-1}\}$ that minimizes the expected total sensitivity-weighted error:

\begin{equation}
\begin{aligned}
    \min_{\{b_l\}} \quad & \mathcal{L}_{\text{total}} = \sum_{l=0}^{L-1} \mathcal{S}_l \cdot \mathcal{E}^{(l)}(b_l) \\
    \text{s.t.} \quad & \sum_{l=0}^{L-1} b_l \le B_{\text{total}}, \quad b_l \in \mathbb{Z}_{\ge 0}.
\end{aligned}
\label{eq:budget_opt}
\end{equation}

Solving Eq. (\ref{eq:budget_opt}) exactly is NP-hard. However, assuming a token's marginal gain is proportional to its layer's intrinsic entropy, we approximate it by allocating the total budget $B_{\text{total}}$ according to a composite importance score $\rho^{(l)}$.

\begin{figure}[t] 
    \centering
    \includegraphics[width=0.7\linewidth]
    {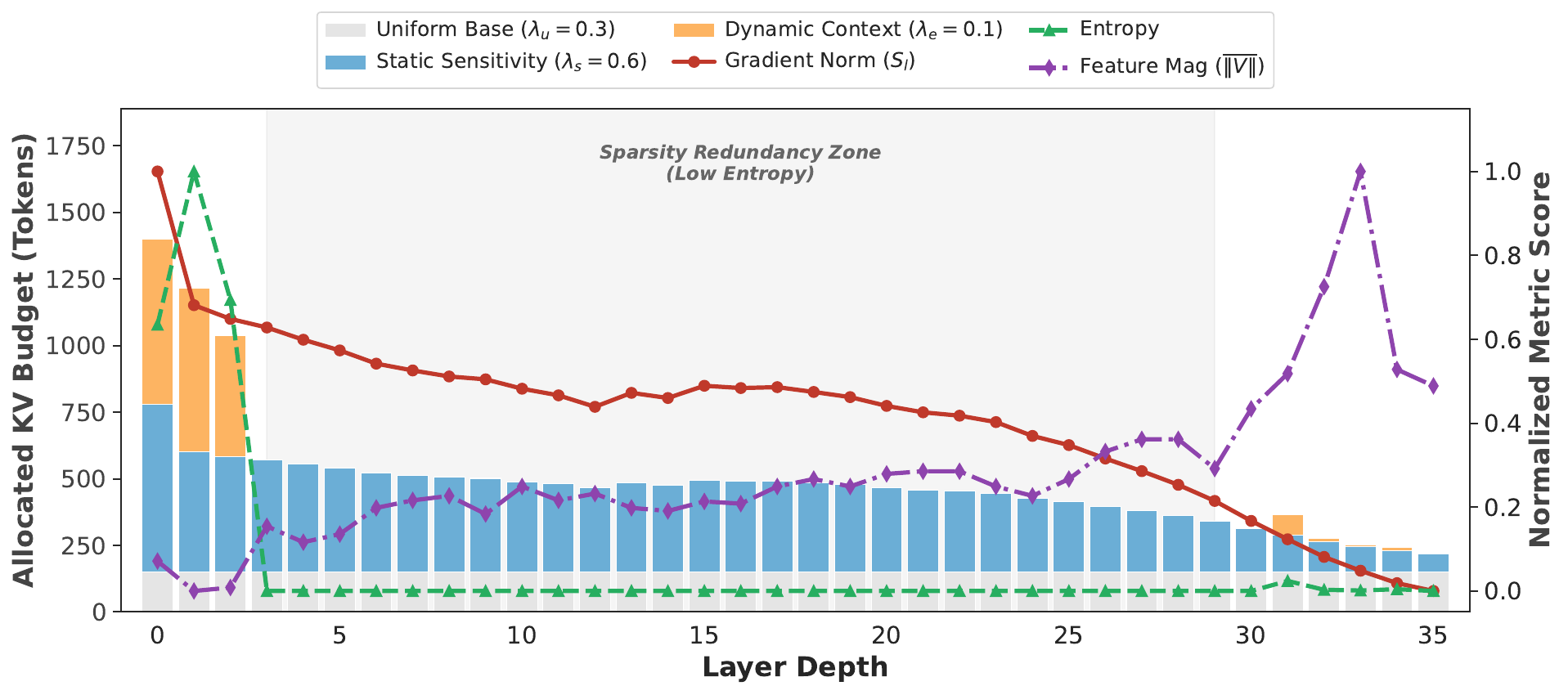}
   
    \caption{{Statistical Visualization on Qwen2.5-VL-3B.} 
Layer-wise budget allocation (stacked bars) is shown under an average budget of 500 KV tokens per layer. Gradient Norm ($S_l$), Entropy ($\mathcal{H}$), and Feature Magnitude ($\overline{\|V\|}$) are normalized for comparison. The strategy effectively assigns higher budgets to shallow layers, while more aggressively pruning deeper layers. See Appendix \ref{app:stat_analysis} for details.
}

    \label{fig:stats}
\end{figure}

\noindent \textbf{Tri-Hybrid Budget Allocation Strategy.} 
We approximate the optimal budget allocation via a linear combination of three complementary metrics. 
As visualized in Figure~\ref{fig:stats}, this strategy yields a {front-heavy budget distribution}: 
more resources are assigned to shallow layers to mitigate error propagation from high $\mathcal{S}_l$ and entropy, while the budget gradually decreases toward deeper layers, enabling increasingly aggressive pruning.
Formally, the budget proportion $\rho^{(l)}$ for the $l$-th layer is defined as:
\begin{equation}
    \rho^{(l)} = \lambda_u \rho_{\text{unif}} + \lambda_s \rho_{\text{static}}^{(l)} + \lambda_e \rho_{\text{dyn}}^{(l)},
\end{equation}
where $\lambda_u, \lambda_s, \lambda_e$ control the trade-off between components. Theoretical derivations are provided in Appendix \ref{app:budget_theory}. The three components are defined as follows:
\begin{itemize}[leftmargin=*, nosep]
    \item {Uniform Regularization ($\rho_{\text{unif}} = 1/L$):} Ensures a minimum budget for all layers to prevent resource starvation 
    and guarantee baseline representational robustness.

    \item {Layer-wise Structural Sensitivity ($\rho_{\text{static}}^{(l)}$):} Derived from Eq.~(\ref{eq:cal}), this term encodes the model's static topology and prioritizes layers with high expected gradients to mitigate cascading amplification of approximation errors.

    \item {Information Density ($\rho_{\text{dyn}}^{(l)}$):} This component aids in reducing the {local approximation error} $\mathcal{E}_l$ by adapting to input-dependent complexity. Following \citet{shen2025lava}, layers with high attention entropy (diffuse information) and large feature magnitudes (strong activation) are identified as resistant to compression. We propose a Log-Norm Weighted Entropy metric:
\begin{equation}
    \mathcal{M}^{(l)}_{\text{dyn}} = \log\big(1 + \mathcal{H}(\mathbf{A}^{(l)})\big) \cdot \overline{\log\big(\|\mathbf{V}^{(l)}\|_2 + \epsilon\big)},
\end{equation}
where $\mathcal{H}(\mathbf{A}^{(l)})$ quantifies attention sparsity. Higher entropy indicates a more uniform and harder-to-compress distribution, while $\|\mathbf{V}^{(l)}\|_2$ reflects activation intensity. By integrating these factors, we preserve layers that are both structurally diverse and functionally important. Finally, we obtain the normalized budget proportion as:
\begin{equation}
    \rho_{\text{dyn}}^{(l)} = \mathcal{M}_{\text{dyn}}^{(l)} \Big/ \sum_{k=0}^{L-1} \mathcal{M}_{\text{dyn}}^{(k)}.
\end{equation}

\end{itemize}

The full budget allocation and pruning procedure is summarized in Algorithm~\ref{alg:kv_compression}.

\begin{algorithm}[h]
\caption{Sensitivity-Aware Budget Allocation}
\label{alg:kv_compression}
\begin{algorithmic}[1]
    \REQUIRE KV Cache states $\{\mathcal{P}^{(l)}\}_{l=1}^L$, Total Budget $B_{\text{total}}$
    \REQUIRE \textbf{Offline Prior:} Structural Sensitivity $S_l = \mathbb{E}[\|\nabla A^{(l)}\|]$
    \ENSURE Compressed Cache $\{\mathcal{P}'^{(l)}\}_{l=1}^L$

    \STATE \textbf{1. Online Metric Evaluation}
    \STATE Generate proxy query $\mathcal{Q}_{\text{proxy}}$ via template
    \FOR{$l = 1$ \textbf{to} $L$ \textbf{in parallel}}
        \STATE $\mathbf{A}^{(l)} \leftarrow \text{Softmax}(\mathcal{Q}_{\text{proxy}} (\mathbf{K}^{(l)})^\top / \sqrt{d})$
        \STATE $\mathbf{s}^{(l)}_{\text{imp}} \leftarrow \text{Mean}(\mathbf{A}^{(l)})$ \quad \textcolor{gray}{// Metric for Eviction}
        
        \STATE \textit{// Estimate compressibility via Entropy}
        \STATE $\mathcal{H}^{(l)} \leftarrow \text{Entropy}(\mathbf{A}^{(l)})$
        \STATE $w^{(l)} \leftarrow \log(1 + \mathcal{H}(\mathbf{A}^{(l)})) \cdot \overline{\log(\|\mathbf{V}^{(l)}\|_2 + \epsilon)}$ \\\textcolor{gray}{// Proxy for $\mathbb{E}[\|\Delta A\|]$}
    \ENDFOR
    
    \STATE \textbf{2. Solving Allocation Problem (Approx.)}
    \STATE Normalize dynamic proxy: $\boldsymbol{\rho}_{\text{dyn}} \leftarrow \mathbf{w} / \|\mathbf{w}\|_1$
    \STATE Normalize static prior: $\boldsymbol{\rho}_{\text{static}} \leftarrow \mathbf{S} / \|\mathbf{S}\|_1$
    \STATE Compute mixed policy:
    \STATE \quad $\boldsymbol{\rho}^* \leftarrow \lambda_u \boldsymbol{\rho}_{\text{unif}} + \lambda_s \boldsymbol{\rho}_{\text{static}} + \lambda_e \boldsymbol{\rho}_{\text{dyn}}$
    \STATE Discretize layer budgets: $\mathbf{b} \leftarrow \lfloor B_{\text{total}} \cdot \boldsymbol{\rho}^* \rfloor$
    
    \STATE \textbf{3. Execution: Layer-wise Pruning}
    \FOR{$l = 1$ \textbf{to} $L$}
        \STATE Select indices $\mathcal{I} \leftarrow \text{TopK}(\mathbf{s}^{(l)}_{\text{imp}}, k=b_l)$
        \STATE Update state: $\mathcal{P}'^{(l)} \leftarrow \mathcal{P}^{(l)}[\mathcal{I}]$
    \ENDFOR
    
    \STATE \textbf{return} $\{\mathcal{P}'^{(l)}\}_{l=1}^L$

\end{algorithmic}
\end{algorithm}

\begin{table*}[t]
\centering
\small
\definecolor{TableHeader}{RGB}{245, 245, 245}
\definecolor{OursColor}{RGB}{236, 246, 255}

\caption{{Performance Comparison on Offline Video Understanding Benchmarks.} 
We evaluate {Video-MemComp} against state-of-the-art compression methods and full-cache baselines. {Avg Budget} indicates the KV cache token limit; \textbf{bold} and \underline{underlined} values mark the best and second-best results, respectively. Entries marked with * denotes reproduced results, others are from original papers. Rows in \colorbox{OursColor}{light blue} show that our method meets or exceeds full-cache performance under strict memory constraints.}

\label{tab:main_result1}

\renewcommand{\arraystretch}{1}

\resizebox{\linewidth}{!}{
  \begin{tabular}{l c c c c c c c c c}
    \toprule
    \rowcolor{TableHeader} 
    \textbf{Method} & \textbf{Size} & \textbf{\# Frames} & \textbf{Avg Budget} & \textbf{EgoSchema} & \textbf{MLVU} & \multicolumn{3}{c}{\textbf{VideoMME}} & \textbf{ALL} \\
    \cmidrule(lr){7-9} 
    \rowcolor{TableHeader} 
    Max Duration & & & \textbf{$|M|$} & 3\,min & 120\,min & Medium & Long & Avg & – \\
    \midrule
    
    GPT4-V & – & 1fps & – & 55.6 & – & – & – & 60.7 & – \\
    GPT4-o & – & 1fps & – & 72.2 & 66.2 & – & – & 77.2 & 71.9 \\
    LongVU & 7B & 1fps & 8K & 67.6 & 65.4 & – & – & 60.6 & 64.5 \\
    LongVU & 3B & 1fps & 8K & 59.1 & 55.9 & – & – & 51.5 & 55.5 \\
    \midrule
    
    Qwen-2-VL* & 7B & 768 & 100K & 65.2 & 65.8 & \textbf{63.9} & \textbf{55.1} & \textbf{63.7} & 64.9 \\
    Qwen-2-VL + {DyCoke}& 7B& 768& 6K& –& 58.0& 55.0& 51.7& 59.8&–\\
    Qwen-2-VL + {InfiniPot-V} & 7B & 768 & 6K & 65.6 & 65.8 & 60.8 & \underline{53.4} & \underline{62.8} & 64.7 \\
    Qwen-2-VL + {StreamMem} & 7B & 0.5/0.2fps & 6K & \underline{67.2} & \underline{65.9} & \underline{62.4} & 52.3 & 62.1 & \underline{65.1} \\
    \rowcolor{OursColor} 
    Qwen-2-VL + \textbf{Video-MemComp} (Ours) & 7B & 768 & 6K & \textbf{68.6} & \textbf{67.1} & 60.2 & 51.6 & 60.8 & \textbf{65.5} \\
    \midrule
    
    {LLaVA-OneVision-7B} & 7B & 32 & 6K & 60.1 & 64.7 & \textbf{57.4} & 46.2 & 56.9 & 60.6 \\
    LLaVA-OneVision-7B + {ReKV} & 7B & 0.5 fps & 6K & 60.7 & \textbf{68.5} & – & – & – & – \\
    LLaVA-OneVision-7B + {LiveVLM} & 7B & 0.5/0.2 fps & - & \textbf{63.0} & 66.3 & 56.4 & \underline{48.8} & 57.3 & 62.2 \\
    LLaVA-OneVision-7B + {StreamMem} & 7B & 0.5/0.2 fps & 6K & \textbf{63.0} & \underline{66.9} & 56.6 & \textbf{50.1} & \textbf{59.4} & \textbf{63.1} \\
    \rowcolor{OursColor}
    LLaVA-OneVision-7B + \textbf{Video-MemComp} (Ours) & 7B & 0.5/0.2 fps & 6K & \textbf{63.0} & 66.5 & \underline{57.2} & 47.3 & \underline{58.0} & \underline{62.5} \\
    \midrule
    
    Qwen-2.5-VL* & 3B & 768 & 100K & \textbf{64.4} & \underline{63.3} & 58.1 & \underline{50.6} & 59.2 & \underline{62.3} \\
    Qwen-2.5-VL + {InfiniPot-V} & 3B & 768 & 6K & 61.8 & 62.1 & – & – & 59.3 & 61.1 \\
    Qwen-2.5-VL + {StreamMem} & 3B & 4.0/0.5 fps & 6K & 62.2 & 62.3 & \textbf{60.1} & 49.1 & \underline{59.5} & 61.3 \\
    \rowcolor{OursColor}
    Qwen-2.5-VL + \textbf{Video-MemComp} (Ours) & 3B & 768 & 6K & \underline{62.8} & \textbf{68.2} & \underline{60.0} & \textbf{53.2} & \textbf{61.2} & \textbf{64.1} \\
    \midrule
    
    InternVL-3.5* & 2B & 8/16/32/64 & 16K & 64.4 & \textbf{64.2} & - & - & \textbf{58.4} & \underline{62.3} \\
    InternVL-3.5 + {InfiniPot-V}* & 2B & 64/100 & 6K & \textbf{67.6} & 58.9 & 46.9 & 41.2 & 48.0 & 58.1 \\
    InternVL-3.5 + {StreamMem}* & 2B & 64/100 & 6K & 58.5 & 59.6 & \underline{51.0} & \underline{43.3} & 51.3 & 56.5 \\
    \rowcolor{OursColor}
    InternVL-3.5 + \textbf{Video-MemComp} (Ours) & 2B & 64/100 & 6K & \underline{65.6} & \underline{63.6} & \textbf{57.4} & \textbf{46.9} & \underline{57.9} & \textbf{62.4} \\
    \bottomrule
  \end{tabular}
}
\label{tab:main_result}
\end{table*}

\section{Experiments}
\subsection{Experimental Setup}

\noindent \textbf{Benchmarks and Evaluation.}
We evaluate Video-MemComp on a comprehensive suite of offline and streaming benchmarks. For offline datasets, we use VideoMME \cite{fu2025video}, MLVU \cite{zhou2025mlvu}, and EgoSchema \cite{mangalam2023egoschema} following the online streaming evaluation protocol. Following prior work \cite{di2025rekv, yang2025streammem, kim2025infinipot}, results are reported on the official development sets. For streaming evaluation, we employ timestamped QA datasets RVS-Ego/Movie \cite{di2025rekv} (using GPT-based evaluation) and the choice-based OVO-Bench \cite{niu2025ovo} to assess temporal reasoning tasks such as backward tracing and real-time perception. Due to space limitations, detailed experimental configurations and results on OVO-Bench are provided in \ref{app:ovo} of the supplementary.




\noindent \textbf{Models.} We evaluate our method across four MLLM backbones: Qwen-2-VL-7B \cite{wang2024qwen2}, Qwen-2.5-VL-3B \cite{bai2025qwen2}, LLaVA-OV-7B \cite{li2024llava}, and InternVL-3.5-2B \cite{wang2025internvl3}. To accommodate architectural variations, we adopt model-specific hyperparameters, including similarity thresholds ($\theta_{\text{inter}}, \theta_{\text{intra}}$) and budget weights ($\lambda_u, \lambda_s, \lambda_e$), as detailed in \ref{app:hyperparams} of the supplementary material.

\subsection{Evaluation Results}

\noindent \textbf{Offline Video Understanding.}
We evaluate Video-MemComp against state-of-the-art compression methods (e.g., InfiniPot-V \cite{kim2025infinipot}, StreamMem \cite{yang2025streammem}, ReKV\cite{di2025rekv}, DyCoke\cite{tao2025dycoke}, LiveVLM\cite{ning2025livevlm}) and full-input baselines across four backbones. As detailed in Table~\ref{tab:main_result1}, our method achieves comparable or superior performance under a restricted 6K KV budget. Notably, on Qwen-2-VL-7B, it scores 65.5, outperforming both InfiniPot-V (64.7) and the full-input baseline (64.9). This efficacy extends to Qwen-2.5-VL-3B (64.1 vs. 62.3) and other backbones (InternVL-3.5-2B, LLaVA-OneVision-7B), demonstrating robust redundancy filtering capabilities.

\noindent \textbf{Streaming Video Understanding.}
We compare with state-of-the-arts~\cite{di2025rekv,zhang2024flash,kim2025infinipot,yang2025streammem,chen2025streamingtom} on RVS-Ego and RVS-Movie (Table~\ref{tab:rvs_benchmarks}). While ReKV achieves 59.0\% via CPU offloading, it incurs high system overhead, and restricting it to GPU memory drops accuracy to 53.3\%. In contrast, our method remains robust within GPU budgets, achieving the highest average accuracy (56.6\%) among efficient methods, surpassing StreamingTOM (55.8\%). Notably, on RVS-Ego, we reach 60.1\%, outperforming StreamingTOM (58.3\%) and InfiniPot-V (57.9\%), which further validates the effectiveness of our sensitivity-aware compression in capturing critical semantic dependencies.

\begin{table}[t]
\centering
\small

\caption{Quantitative Results on RVS Streaming Benchmarks.
We compare our method with state-of-the-art streaming baselines, grouped into \textit{Reference (Top)} using CPU offloading for infinite history, and \textit{Memory-Constrained (Bottom)} within GPU limits. 
{Video-MemComp} achieves the highest average accuracy among memory-constrained methods, especially in the egocentric domain.}

\label{tab:rvs_benchmarks}

    \begin{tabular}{lcccccc}
    \toprule
    & \multicolumn{2}{c}{RVS-Ego} & \multicolumn{2}{c}{RVS-Movie} & \multicolumn{2}{c}{Avg} \\
    \cmidrule(lr){2-3} \cmidrule(lr){4-5} \cmidrule(lr){6-7}
    Method & Acc (\%) & Score & Acc (\%) & Score & Acc (\%) & Score \\
    \midrule
    ReKV  & 63.7 & 4.0 & 54.4 & 3.6 & 59.0 & 3.8 \\
    \midrule
    ReKV w/o off.  & 55.8 & 3.3 & 50.8 & 3.4 & 53.3 & 3.4 \\
    Flash-VStream  & 57.0 & \textbf{4.0} & \underline{53.1} & 3.3 & 55.0 & \underline{3.6}\\
    InfiniPot-V  & 57.9 & 3.5 & 51.4 & \textbf{3.5} & 54.6 & 3.5 \\
    StreamMem  & 57.6 & 3.8 & 52.7 & 3.4 & 55.2 & \underline{3.6} \\
    StreamingTOM  & \underline{58.3} & \underline{3.9} & \textbf{53.2} & \textbf{3.5} & \underline{55.8} & \textbf{3.7} \\
    \textbf{Video-MemComp} (Ours) & \textbf{60.1} & 3.3 & \underline{53.1} & 3.1 & \textbf{56.6} & 3.2 \\
    \bottomrule
    \end{tabular}


\end{table}

\noindent \textbf{Efficiency Analysis.}
To further verify the potential of our method for deployment on edge devices, we conducted a detailed efficiency evaluation on an NVIDIA RTX A6000 based on the LLaVA-OneVision-7B model. As shown in Table~\ref{tab:efficiency_analysis}, we focus on the comparison with the streaming method ReKV. Although ReKV achieves higher throughput, our method demonstrates clear advantages in GPU memory usage and interaction latency, which are key metrics determining user experience.

\begin{table}[h]
\centering
\footnotesize
\caption{System efficiency on LLaVA-OneVision-7B. Metrics include peak memory, visual encoder latency per frame, and time to first token (TTFT). Video-MemComp reduces memory and latency compared to retrieval-based baselines (ReKV), enabling OOM-free consumer-grade inference.}
\label{tab:efficiency_analysis}
\setlength{\tabcolsep}{8pt} 

\begin{tabular}{l c c c c}
\toprule
{Method} & {Peak Memory} & {Video Enc.} & {TTFT} & {Throughput} \\
 & (GB) $\downarrow$ & (ms/frame) $\downarrow$ & (s) $\downarrow$ & (tokens/s) $\uparrow$ \\
\midrule
LLaVA-OV (Baseline) & OOM& 220.48& 0.0600& 6.45\\
 ReKV & 22.95& 82.75& 0.0846&35.61\\
 ReKV w/o off.& 22.95& 90.49& 0.0663&\textbf{36.93}\\
\textbf{Video-MemComp} (Ours) & \textbf{16.74}& \textbf{64.92}& \textbf{0.0302}& 27.58\\
\bottomrule
\end{tabular}

\end{table}

\subsection{Ablation Study}

We conduct comprehensive ablation studies to disentangle the contributions of the two theoretically orthogonal components derived in Eq.~(\ref{eq:taylor_proof}): 
{(i) minimizing local approximation error} $\mathcal{E}_l$ via intra-layer compression, and 
{(ii) managing layer-wise structural sensitivity} $\mathcal{S}_l$ via inter-layer budget allocation.
Unless otherwise specified, all experiments are conducted on {VideoMME} (streaming at 4.0fps / 0.5fps) and {MLVU} (full 768-frame) using {Qwen2.5-VL-3B} .

\begin{figure}[ht]
\vskip 0.2in
\begin{center}
\centerline{\includegraphics[width=0.55\linewidth]{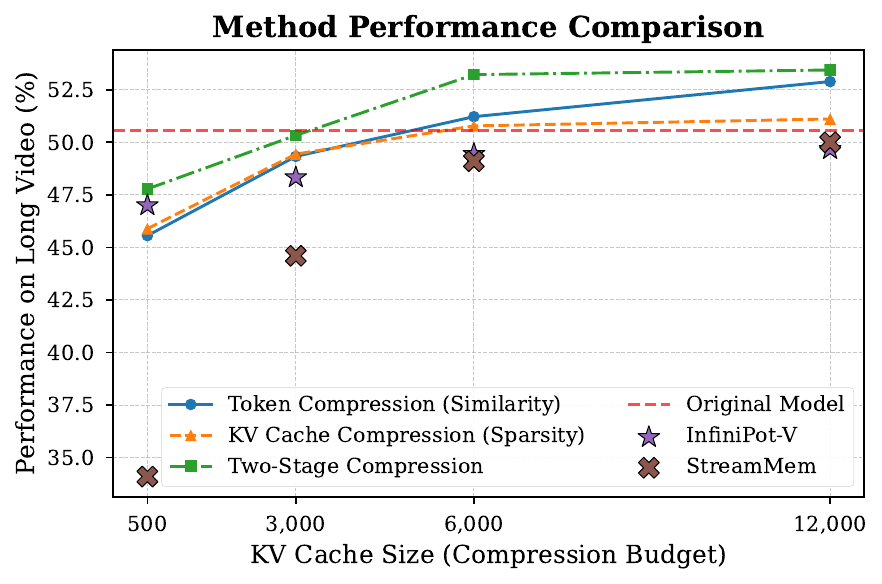}} 

\caption{Accuracy-Memory trade-offs on Qwen2.5-VL-3B (VideoMME Long Video).
Token Compression (Blue) excels in high-budget regimes, notably surpassing the dense baseline by mitigating attention dilution. Conversely, KV Compression (Orange) demonstrates robustness under strict memory constraints. Our Hybrid Method (Green) achieves the \emph{optimal Pareto frontier} across all budgets, balancing redundancy removal with attention preservation. See \ref{app:hyperparams} in the supplementary material for details.
}
\label{compression_performance}
\end{center}
\vskip -0.2in
\end{figure}

\noindent \textbf{Hybrid Intra-layer Compression vs. Single-Mechanism Baselines.}
We validate the necessity of our {hybrid intra-layer strategy}, which combines {Pre-RoPE token aggregation} with {sparsity-based KV pruning} to jointly minimize the local approximation error $\mathcal{E}_l$. To isolate intra-layer effects, we employ a \emph{uniform layer-wise budget} (disabling $\mathcal{S}_l$-aware allocation). As shown in Fig.\ref{compression_performance}, this hybrid design consistently dominates single-mechanism baselines, achieving superior accuracy-memory trade-offs by simultaneously addressing representation redundancy and attention sparsity.

\noindent \textbf{Effect of Intra-frame Token Aggregation on $\mathcal{E}_l$.}
We analyze intra-frame token aggregation, which targets spatial redundancy and directly reduces the local approximation error $\mathcal{E}_l$. 
Table~\ref{tab:ablation_intra} compares an inter-frame-only baseline with full spatio-temporal aggregation. 
While inter-frame filtering removes static temporal redundancy, intra-frame aggregation densifies spatial semantics within each frame, consistently improving long-context benchmarks (e.g., MLVU from 67.68 to 68.19, and gains on VideoMME Medium and Long). 
Minor drops on short clips indicate a benign trade-off, as spatial fusion is less critical with limited temporal context. 
Overall, intra-frame aggregation effectively lowers $\mathcal{E}_l$ under tight memory constraints without sacrificing essential semantics.


\definecolor{posgreen}{RGB}{0, 150, 0} 
\definecolor{negred}{RGB}{150, 0, 0}   

\begin{table}[h]
\small
\centering


\caption{Ablation study on intra-frame token compression. The inter-frame only baseline applies temporal filtering ($\theta_{\text{inter}}=0.3$) with standard KV pruning. Adding intra-frame compression improves performance on longer contexts by densifying spatial semantics, with a minor trade-off on short clips.}

\label{tab:ablation_intra}
\setlength{\tabcolsep}{3.5pt}

\begin{tabular}{l ccccc} 
\toprule
& \multicolumn{4}{c}{{VideoMME}} & \\ 
\cmidrule(lr){2-5} 
{Method} & S & M & L & Avg & {MLVU} \\ 
\midrule
Inter-frame Only & \textbf{70.7} & 58.6 & 52.3 & 60.5 & 67.68 \\ 
Inter + {Intra-frame} & 70.3 & \textbf{59.3} & \textbf{52.8} & \textbf{60.8} & \textbf{68.19} \\ 
\midrule
$\Delta$ & \textcolor{gray}{-0.4} & \textcolor{posgreen}{+0.7} & \textcolor{posgreen}{+0.5} & \textcolor{posgreen}{+0.3} & \textcolor{posgreen}{+0.51} \\ 
\bottomrule
\end{tabular}

\end{table}

\noindent \textbf{Sensitivity-Aware vs. Uniform Budget Allocation.}
We compare sensitivity-aware dynamic budget allocation with static uniform allocation to evaluate the role of layer-wise structural sensitivity $\mathcal{S}_l$.
All variants use identical intra-layer compression, isolating inter-layer scheduling effects.
As shown in Table~\ref{tab:ablation_dynamic}, the dynamic strategy consistently outperforms uniform allocation across similarity thresholds, with the largest gains on long-video and aggregate metrics.
These results show that prioritizing structurally sensitive layers suppresses error amplification, empirically confirming that $\mathcal{S}_l$ is a global error amplifier requiring explicit control.

\begin{table}[h]
\centering
\small

\caption{Ablation study on static vs. dynamic budget allocation on VideoMME. Our sensitivity-aware dynamic strategy consistently improves aggregate performance across similarity thresholds ($\tau=0.75, 0.90$), validating the benefit of prioritizing structurally sensitive and high-entropy layers over a uniform distribution.}

\label{tab:ablation_dynamic}

\begin{tabular}{c l c c c c}
\toprule
{Threshold} & {Strategy} & {Short} & {Medium} & {Long} & {All} \\ 
\midrule
\multirow{2}{*}{0.75} & Static & 70.0 & \textbf{60.3} & 53.2 & 61.17 \\ 
 & {Dynamic} & \textbf{70.5} & 60.0 & \textbf{53.2} & \textbf{61.23} \\ 
\midrule
\multirow{2}{*}{0.90} & Static & 70.0 & \textbf{60.7} & 52.3 & 61.00 \\ 
 & {Dynamic} & \textbf{71.0} & 59.9 & \textbf{53.0} & \textbf{61.53} \\ 
\bottomrule
\end{tabular}

\end{table}

\noindent \textbf{Ablation of Budget Allocation Weights.}
We study the effect of the tri-hybrid budget allocation weights (uniform prior $\lambda_u$, structural sensitivity $\lambda_s$, and dynamic information density $\lambda_e$) on OVO-Bench using Qwen2.5-VL-7B, with fixed spatio-temporal thresholds.
Allocations dominated by structural sensitivity ($\lambda_s$) consistently outperform uniform-heavy configurations, confirming the effectiveness of offline gradient-based calibration.
Meanwhile, introducing a small dynamic component ($\lambda_e \approx 0.025$) yields the global optimum, as it adaptively mitigates input-dependent approximation error without destabilizing the allocation.
Excessive reliance on dynamic metrics leads to performance degradation due to high variance in online activations.

\begin{figure}[h]
    \centering
    \includegraphics[width=0.55\linewidth]{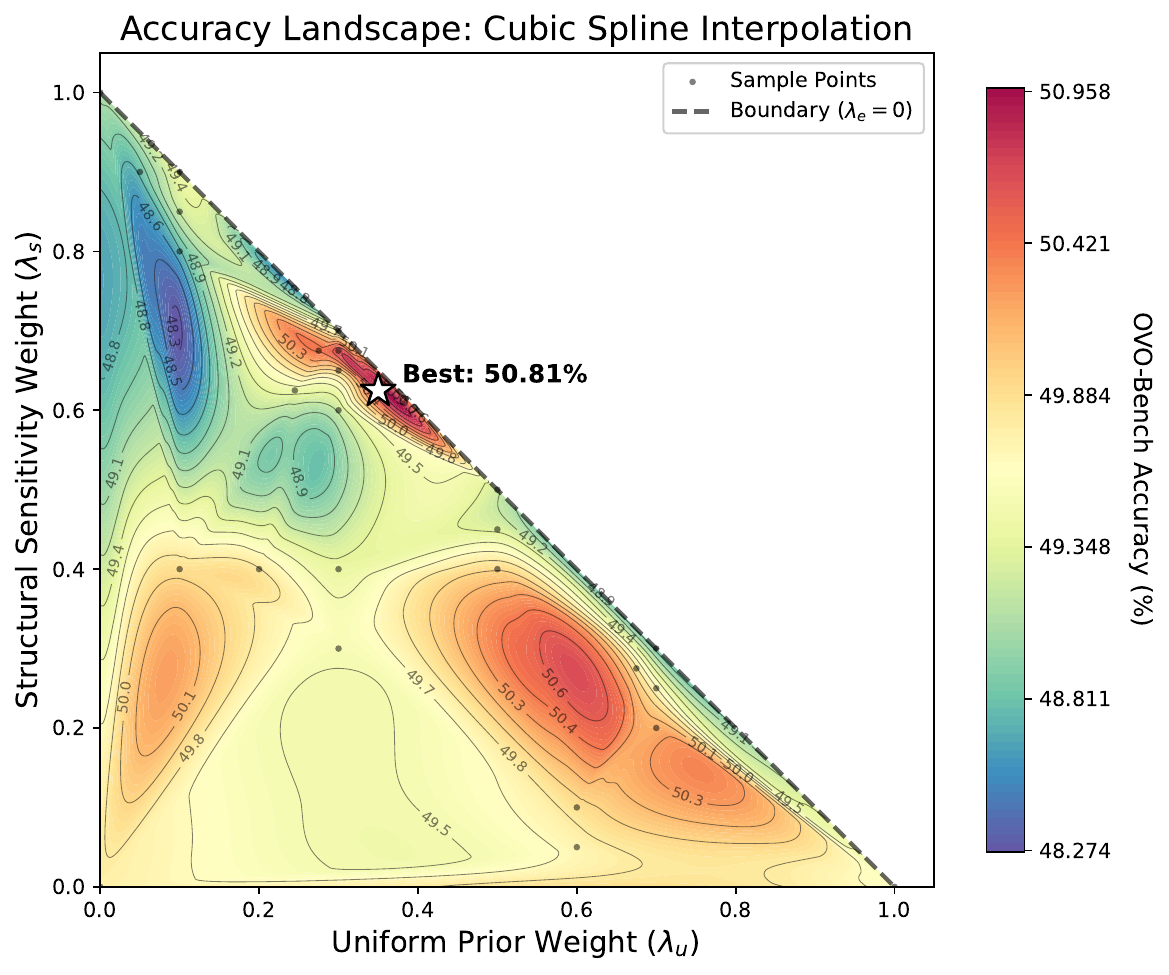}
    \caption{Hyperparameter landscape of budget allocation. OVO-Bench accuracy is shown as a contour over uniform prior ($\lambda_u$) and structural sensitivity ($\lambda_s$); $\lambda_e = 1-\lambda_u-\lambda_s$ represents the dynamic entropy weight. The global optimum(star), revealing that a dominant structural prior combined with subtle dynamic adjustment yields superior performance compared to uniform allocation. See \ref{app:buget_ablation} in the supplementary material for details.
}

    \label{fig:ablation_contour}
\end{figure}

\begin{figure}[h]
    \centering
    \includegraphics[width=0.55\linewidth]{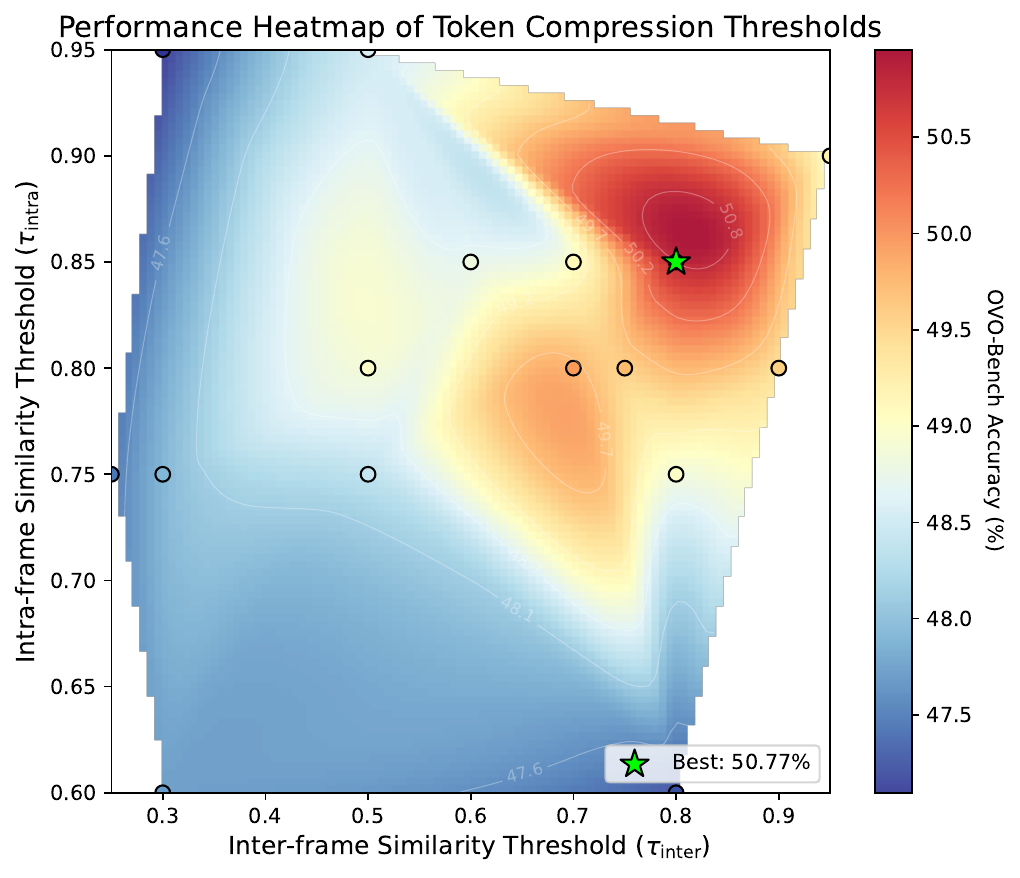}
    \caption{{Joint Impact of Spatio-Temporal Compression Thresholds.} 
    The heatmap plots accuracy across varying Inter-frame ($\theta_{\text{inter}}$) and Intra-frame ($\theta_{\text{intra}}$) similarity thresholds. The landscape reveals a convex optimization zone, where the peak represents the optimal trade-off between aggressive redundancy filtering and semantic preservation. Notably, overly aggressive temporal pruning (low $\theta_{\text{inter}}$) causes irreversible performance degradation. See \ref{app:sim_ablation} in the supplementary material for details.
}
    \label{fig:sim_heatmap}
\end{figure}

\noindent \textbf{Sensitivity to Spatio-temporal Similarity Thresholds.}
Finally, we analyze the interaction between inter-frame ($\theta_{\text{inter}}$) and intra-frame ($\theta_{\text{intra}}$) similarity thresholds, which jointly determine the granularity of redundancy removal.
Figure~\ref{fig:sim_heatmap} shows a convex performance landscape. Aggressive temporal pruning ($\theta_{\text{inter}} < 0.5$) causes irreversible accuracy collapse, highlighting that temporal continuity is a prerequisite for streaming video understanding.
Performance peaks at $\theta_{\text{inter}}=0.8$ and $\theta_{\text{intra}}=0.85$, where static redundancy is effectively removed without sacrificing fine-grained visual cues.
Overly conservative thresholds, by contrast, retain excessive redundancy and exhaust the limited KV budget, increasing $\mathcal{E}_l$ and degrading overall performance.

\section{Conclusion}

In this work, we present \textbf{Video-MemComp}, a training-free, model-agnostic, and structure-aware hybrid framework for infinite streaming video understanding under a strict $O(1)$ memory budget.
Moving beyond heuristic pruning, we revisit KV cache compression through the lens of error propagation.
Using a first-order Taylor expansion as an analytical surrogate, we decompose the global output error into two orthogonal components: \textit{Local Approximation Error} ($\mathcal{E}_l$) and \textit{Layer-wise Structural Sensitivity} ($\mathcal{S}_l$), revealing the mechanism of non-uniform error propagation in MLLMs.
Guided by this insight, Video-MemComp reduces $\mathcal{E}_l$ via Pre-RoPE aggregation that bypasses RoPE-induced geometric constraints, and regulates $\mathcal{S}_l$ through gradient-guided dynamic budget allocation.
Extensive experiments demonstrate a counter-intuitive “less is more” effect: by distilling high-density information, our method matches or surpasses full-cache baselines with minimal memory.
This work offers a scalable solution for long-form video understanding on resource-constrained devices and provides a principled perspective on memory optimization.

\paragraph{Limitations and Future Work.}
Our Taylor-based decomposition is heuristic and may break down under extreme compression where local perturbation assumptions no longer hold.
The structural sensitivity prior relies on offline calibration and may degrade under distribution shifts, while Pre-RoPE aggregation introduces moderate overhead.
Future work will explore validating higher-order approximations, developing online or adaptive sensitivity estimation, and designing more efficient aggregation for ultra-long streams.

\begin{flushleft}
\bibliographystyle{plainnat}
\bibliography{citation}
\end{flushleft}

\clearpage
\appendix

\newpage
\section{Appendix Roadmap}

This appendix is organized as follows: (i) \cref{app:theory_details} formulates the streaming video understanding problem and derives the first-order Taylor error decomposition, serving as our theoretical foundation; (ii) \cref{app:theoretical_derivation} provides the theoretical justification and implementation details for our hybrid compression scheme to minimize local approximation error; (iii) \cref{app:budget_theory} presents the rigorous derivation of our sensitivity-aware budget allocation for managing layer-wise structural sensitivity; and (iv) \cref{app:Hyperparameter,app:ovo} report supplementary experimental configurations and extended evaluations.

\section{Problem Formulation and Taylor-based Error Decomposition}
\label{app:theory_details}

\subsection{Problem Formulation}

Consider a streaming video MLLM processing a sequence of visual tokens.
Let the generation process at time step $t$ be represented as a function $G$ mapping the historical Key-Value (KV) cache to the output \textbf{hidden states or logits} (before discrete sampling).
For the $l$-th layer, let the accumulated cache be denoted as $\mathcal{H}^{(l)} = \{ (\mathbf{k}_i, \mathbf{v}_i) \}_{i=1}^{t-1}$, where $\mathbf{k}_i, \mathbf{v}_i \in \mathbb{R}^{d}$ represent the Key and Value vectors.
The global model output $\mathcal{O}$ depends on the aggregate state of all $L$ layers, denoted as $\mathbb{H} = \{ \mathcal{H}^{(0)}, \dots, \mathcal{H}^{(L-1)} \}$, such that $\mathcal{O} = G(\mathbb{H})$.

Our objective is to identify a compression operator $\mathcal{T}$ that maps the full cache $\mathbb{H}$ to a sparse, memory-efficient set $\mathbb{H}' = \mathcal{T}(\mathbb{H})$ under a strict global memory budget $B_{\text{total}}$. This is formulated as a constrained optimization problem:
\begin{equation}
\label{eq:appendix_opt}
    \min_{\mathcal{T}} \|\Delta \mathcal{O}\| = \| G(\mathcal{T}(\mathbb{H})) - G(\mathbb{H}) \|, \quad \text{s.t.} \sum_{l=0}^{L-1} |\mathcal{H}'^{(l)}| \le B_{\text{total}}.
\end{equation}

\subsection{Error Analysis via First-Order Taylor Approximation}

Directly minimizing Eq.~(\ref{eq:appendix_opt}) is intractable due to the high-dimensional non-linearity of $G$. To derive a feasible surrogate, we analyze error propagation through the attention mechanism. We denote the attention output of layer $l$ as $A^{(l)} = \text{Attn}(Q^{(l)}, \mathcal{H}^{(l)})$. 
Compression introduces a local deviation $\Delta \mathbf{A}^{(l)} = \text{Attn}(\mathbf{Q}^{(l)}, \mathcal{T}(\mathcal{H}^{(l)})) - \mathbf{A}^{(l)}$.

By applying a first-order Taylor expansion to $G$ with respect to the attention outputs $\{\mathbf{A}^{(l)}\}_{l=0}^{L-1}$, we decompose the global output error as follows:
\begin{equation}
\label{eq:taylor_bound_new}
    \|\Delta \mathcal{O}\| \approx \left\| \sum_{l=0}^{L-1} \frac{\partial G}{\partial \mathbf{A}^{(l)}} \Delta \mathbf{A}^{(l)} \right\| 
    \le \sum_{l=0}^{L-1} \underbrace{\left\| \frac{\partial G}{\partial \mathbf{A}^{(l)}} \right\|}_{\mathcal{S}_l} \cdot \underbrace{\left\| \Delta \mathbf{A}^{(l)} \right\|}_{\mathcal{E}_l}.
\end{equation}

\textit{Remark: While KV cache pruning introduces discrete and potentially non-infinitesimal perturbations, the first-order Taylor expansion serves as a mathematically principled \textbf{proxy} to disentangle the error sources and inform our budget allocation strategy.}

Eq.~(\ref{eq:taylor_bound_new}) characterizes two distinct error factors:
\begin{itemize}[leftmargin=*, topsep=4pt, itemsep=2pt]
    \item \textbf{Layer-wise Structural Sensitivity ($\mathcal{S}_l$):} This term quantifies the gradient-based amplification factor. High $\mathcal{S}_l$ (typically in shallow layers) indicates that local errors propagate strongly to the final output. This motivates our \textbf{Dynamic Budget Allocation}, prioritizing layers with higher sensitivity.
    
    \item \textbf{Local Approximation Error ($\mathcal{E}_l$):} This term measures information loss within layer $l$. To minimize $\mathcal{E}_l$, we propose the \textbf{Pre-RoPE Aggregation} strategy, which eliminates redundancy while preserving attention mass via logit bias.
\end{itemize}

\subsection{Minimizing Local Error \texorpdfstring{$\mathcal{E}_l$}{E_l}}
\label{sec:minimizing_local_error}

To minimize $\mathcal{E}_l$, we analyze the standard attention formulation. For a query $\mathbf{Q}$, the attention output is:
\begin{equation}
    \mathbf{A} = \sum_{i=1}^{T} p_i \mathbf{v}_i, \quad \text{where } p_i = \frac{\exp(\mathbf{Q}^\top \mathbf{k}_i)}{\sum_{j=1}^{T} \exp(\mathbf{Q}^\top \mathbf{k}_j)}
\end{equation}
Here, $p_i$ represents the attention score of the $i$-th token. Video data redundancy manifests in two orthogonal forms: Similarity and Sparsity. We illustrate how our hybrid strategy minimizes $\mathcal{E}_l$ for each case.

\subsubsection{Case 1: Similarity Redundancy (Minimization via Aggregation)}
\label{app:case1}
\textbf{Premise:} There exists a subset of tokens $\mathcal{I}_{\text{sim}} \subset \{1, \dots, T\}$ that are semantically similar, i.e., $\forall i \in \mathcal{I}_{\text{sim}}, \mathbf{k}_i \approx \mathbf{k}_c, \mathbf{v}_i \approx \mathbf{v}_c$, where $c$ is a representative centroid.

If we naively merge these $N = |\mathcal{I}_{\text{sim}}|$ tokens into one without adjustment, the total attention mass decreases (from a sum of $N$ terms to 1), causing a large shift in $\mathbf{A}$.
To minimize this error, we propose \textbf{Attention-Conserving Aggregation}. 
Let $s = \mathbf{Q}^\top \mathbf{k}_c$ be the logit of the representative. The contribution of the group to the weighted sum was approximately $\sum_{i \in \mathcal{I}_{\text{sim}}} e^s \mathbf{v}_c = N e^s \mathbf{v}_c$.
After merging into a single token $\mathbf{t}_{\text{merged}}$, its new logit $s'$ must satisfy:
\begin{equation}
\label{eq:logit_bias}
    e^{s'} \approx N e^s \implies s' = s + \log(N)
\end{equation}
Adding a $\log(N)$ Logit Bias (generalized as effective count $f_i$) ensures the aggregated token preserves the group's total attention mass. Thus, this subset's error contribution approaches zero:
\begin{equation}
    \|\Delta \mathbf{A}_{\text{sim}}\| = \left\| \left(\sum_{i \in \mathcal{I}_{\text{sim}}} p_i \mathbf{v}_i\right) - p_{\text{merged}} \mathbf{v}_c \right\| \approx 0
\end{equation}
\textit{Remark:} To satisfy the assumption $\mathbf{k}_i \approx \mathbf{k}_c$, this merging must be performed in the \textbf{Pre-RoPE} space to avoid the geometric barrier of rotation.

\subsubsection{Case 2: Sparsity Redundancy (Minimization via Pruning)}

\textbf{Pre-mise:} There exists a subset of tokens $\mathcal{I}_{\text{sparse}}$ where attention scores are negligible, i.e., $\forall j \in \mathcal{I}_{\text{sparse}}, p_j \approx 0$.

For these tokens, aggregation is computationally wasteful. We employ KV Pruning. The error source here is the \textbf{Renormalization} of the softmax denominator (partition function $Z$). 
Let the pruned probability mass be $\epsilon = \sum_{j \in \mathcal{I}_{\text{sparse}}} p_j$. The new attention weight $p'_i$ for retained tokens becomes:
\begin{equation}
    p'_i = \frac{e^{s_i}}{Z - Z_{\text{pruned}}} = \frac{e^{s_i}}{Z(1 - \epsilon)} \approx p_i (1 + \epsilon)
\end{equation}
The resulting approximation error in $\mathbf{A}$ is bounded by the total pruned mass:
\begin{equation}
    \|\Delta \mathbf{A}_{\text{prune}}\| = \left\| \sum_{i \notin \mathcal{I}_{\text{sparse}}} (p'_i - p_i) \mathbf{v}_i - \sum_{j \in \mathcal{I}_{\text{sparse}}} p_j \mathbf{v}_j \right\| \approx O(\epsilon)
\end{equation}
Since our method identifies $\mathcal{I}_{\text{sparse}}$ using a proxy query that explicitly minimizes $\epsilon$ (removing only the least important tokens), the error $\|\Delta \mathbf{A}_{\text{prune}}\|$ is strictly bounded.

\textbf{Conclusion:} By applying \textit{Logit Bias Aggregation} for similarity (Case 1) and \textit{Importance-Guided Pruning} for sparsity (Case 2), we minimize both components of $\|\Delta \mathbf{A}^{(l)}\|$, thereby minimizing the total error $\|\Delta \mathcal{O}\|$ as defined in Eq.~\ref{eq:taylor_bound_new}.

\section{Minimizing Local Approximation Error: Theoretical Justification and Hybrid Compression Implementation}
\label{app:theoretical_derivation}

This appendix provides the theoretical basis and implementation details for the proposed Structural-Aware Hybrid Compression framework. The theoretical analysis covers three core dimensions: (1) Proving that \textbf{cosine similarity} is a \textbf{theoretically grounded} metric for token redundancy under generalized attention mechanisms involving linear projections and Rotary Positional Embeddings ; (2) Elucidating the complementary mechanisms of \textbf{Inter-Frame Filtering} (implicit temporal high-pass filtering) and \textbf{Pre-RoPE Intra-Frame Aggregation} (energy-conserving reconstruction); (3) Modeling adaptive Token Compression as an optimal quantization problem with \textbf{fixed distortion and variable rate} from an information theoretic perspective. Finally, we provide detailed algorithmic implementation steps.

\subsection{Optimality Proof of Cosine Similarity Under Generalized Attention Constraints}
\label{app:metric_optimality}

In Multimodal Large Language Models (MLLMs), determining whether two input tokens $\mathbf{x}_i$ and $\mathbf{x}_j$ are redundant is essentially equivalent to requiring that they produce consistent attention Logit responses to any potential query vector $\mathbf{q}$ after being mapped to the Key space via parameter projection and positional encoding.

\subsubsection{Formalization of Generalized Response Error}

\paragraph{Notations:}
\begin{itemize}[leftmargin=*, nosep]
    \item \textbf{Input Features:} $\mathbf{x}_i, \mathbf{x}_j \in \mathbb{R}^d$. Given the prevalence of normalization layers (LayerNorm) in LLMs, we assume feature vectors lie on a hypersphere, i.e., $\|\mathbf{x}\| \approx C$.
    \item \textbf{Linear Projection:} $\mathbf{W}_k \in \mathbb{R}^{d \times d}$, mapping input features to the Key space.
    \item \textbf{Rotary Positional Embedding (RoPE):} $\mathcal{R}_p \in \mathbb{R}^{d \times d}$ is the orthogonal rotation matrix at position $p$, satisfying $\mathcal{R}_p^\top \mathcal{R}_p = \mathbf{I}$.
    \item \textbf{Arbitrary Query:} $\mathbf{q} \in \mathbb{R}^d$, representing any potential Attention Query generated within the network.
\end{itemize}
The true attention response $S_i$ of token $\mathbf{x}_i$ to query $\mathbf{q}$ is defined as:
\begin{equation}
    S_i(\mathbf{q}) = \mathbf{q}^\top \mathbf{k}_i = \mathbf{q}^\top \mathcal{R}_{p_i} (\mathbf{W}_k \mathbf{x}_i)
\end{equation}

\subsubsection{Local Equivariance and Effective Back-Projection}
We focus on local redundancy (e.g., temporally adjacent frames or spatially neighboring patches). In this context, assuming $\mathbf{x}_i$ and $\mathbf{x}_j$ possess spatio-temporal locality, i.e., $p_i \approx p_j$. Leveraging the continuity of RoPE, we can apply a local isometric approximation: $\mathcal{R}_{p_i} \approx \mathcal{R}_{p_j} \triangleq \mathcal{R}$.
Thus, the response error $\Delta(\mathbf{q})$ can be expressed as:
\begin{equation}
    \Delta(\mathbf{q}) = S_i(\mathbf{q}) - S_j(\mathbf{q}) \approx \mathbf{q}^\top \mathcal{R} \mathbf{W}_k (\mathbf{x}_i - \mathbf{x}_j)
\end{equation}
To trace the error analysis back to the input space $\mathbf{x}$, we rewrite the equation in inner product form:
\begin{equation}
    \Delta(\mathbf{q}) = (\underbrace{\mathbf{W}_k^\top \mathcal{R}^\top \mathbf{q}}_{\tilde{\mathbf{q}}})^\top (\mathbf{x}_i - \mathbf{x}_j) = \tilde{\mathbf{q}}^\top \mathbf{x}_i - \tilde{\mathbf{q}}^\top \mathbf{x}_j
\end{equation}
where we define $\tilde{\mathbf{q}}$ as the \textbf{Effective Back-projected Query}.
\textbf{Physical Interpretation:} Since $\mathbf{W}_k$ is typically full-rank and $\mathcal{R}$ is orthogonal, this transformation is bijective. As $\mathbf{q}$ traverses the $\mathbb{R}^d$ space, $\tilde{\mathbf{q}}$ also traverses the entire feature space. Therefore, maintaining consistent response for arbitrary $\mathbf{q}$ is equivalent to maintaining consistent response for arbitrary $\tilde{\mathbf{q}}$.

\subsubsection{Worst-Case Error Bound Analysis}
The optimization objective can be reformulated as minimizing the discrepancy between $\tilde{\mathbf{q}}^\top \mathbf{x}_i$ and $\tilde{\mathbf{q}}^\top \mathbf{x}_j$. Consider the 2D subspace spanned by $\mathbf{x}_i, \mathbf{x}_j$, and $\tilde{\mathbf{q}}$. Let $\theta$ denote the angle between $\mathbf{x}_i$ and $\mathbf{x}_j$, and let $\phi$ be the azimuth of $\tilde{\mathbf{q}}$ relative to their bisector. Using trigonometric identities, the response error $\Delta$ can be expanded as:
\begin{equation}
\begin{aligned}
    \Delta(\phi, \theta) &= \|\tilde{\mathbf{q}}\| C \left[ \cos\left(\phi - \frac{\theta}{2}\right) - \cos\left(\phi + \frac{\theta}{2}\right) \right] \\
    &= 2\|\tilde{\mathbf{q}}\| C \sin(\phi) \sin\left(\frac{\theta}{2}\right).
\end{aligned}
\end{equation}

In streaming inference, the future query $\mathbf{q}$ is stochastic and unknown. To ensure system robustness, we consider the \textit{worst-case} scenario where $\tilde{\mathbf{q}}$ is orthogonal to the bisector (i.e., $\sin(\phi)=1$). The error upper bound is then given by:
\begin{equation}
    |\Delta|_{\max} \propto \left| \sin\left(\frac{\theta}{2}\right) \right|.
\end{equation}
Minimizing this upper bound (i.e., $\Delta \to 0$) necessitates $\theta \to 0$, which is equivalent to:
\begin{equation}
    \cos\theta \to 1 \iff \text{CosineSim}(\mathbf{x}_i, \mathbf{x}_j) \to 1.
\end{equation}

\textbf{Conclusion.} Even under linear projection $\mathbf{W}_k$ and RoPE rotation $\mathcal{R}$, phase consistency (i.e., cosine similarity) in the input space remains a theoretically rigorous condition for guaranteeing attention response consistency, provided that local positional approximation holds and feature magnitudes are balanced. This justifies the use of cosine similarity as a robust compression metric.
\subsection{Mechanisms of the Dual Compression Strategy}
\label{app:dual_strategy}

Addressing different redundancy types in video, we design complementary dual strategies: Drop-without-Repair for temporal redundancy and Merge-with-Repair for spatial redundancy.

\subsubsection{Inter-Frame: Implicit Temporal High-Pass Filtering}
\label{app:inter_frame_logic}

Let $\mathcal{K}_{t-1}$ be the historical cache and $\mathbf{V}_t$ be the token set of the current frame. We partition $\mathbf{V}_t$ into a static set $\mathcal{S}$ (highly similar to history) and a dynamic set $\mathcal{D}$ (containing new information).

\textbf{Mechanistic Interpretation:}
The \textbf{Marginal Information Gain} of retaining $\mathcal{S}$ is near zero because $\mathcal{K}_{t-1}$ already contains equivalent semantic indices.
By directly discarding $\mathcal{S}$ without re-weighting, we effectively introduce an \textbf{Implicit Attention Re-weighting} mechanism. Consider the Softmax distribution for query $\mathbf{q}$:

\begin{equation}
    \text{Attn}(\mathbf{q}, \mathcal{K}_{\text{all}}) = \frac{\sum_{\mathbf{k} \in \mathcal{D}} e^{\mathbf{q}\cdot\mathbf{k}} \mathbf{v}_{\mathbf{k}} + \sum_{\mathbf{k} \in \mathcal{K}_{t-1}} e^{\mathbf{q}\cdot\mathbf{k}} \mathbf{v}_{\mathbf{k}}}{\mathcal{Z}_{\text{dynamic}} + \mathcal{Z}_{\text{history}} + \underbrace{\mathcal{Z}_{\text{static\_current}}}_{\text{Removed}}}
\end{equation}
Removing the denominator term $\mathcal{Z}_{\text{static\_current}}$ effectively amplifies the relative probability mass of $\mathcal{D}$ (new changes) and $\mathcal{K}_{t-1}$ (valid history).

\textbf{Conclusion:} This operation functionally acts as a \textbf{Temporal High-Pass Filter}, forcing the model to concentrate its limited attention budget on high-frequency change regions, thereby enhancing sensitivity to dynamic events.

\subsubsection{Intra-Frame: Energy Conservation via Logit Bias}
\label{app:intra_frame_logic}

Spatial redundancy (e.g., large background areas) represents low-frequency context that must be preserved. Simply merging tokens without correction leads to the \textbf{Collapse} of total attention energy.
Assume $f_i = |\Omega_i|$ tokens in set $\Omega_i$ are merged into a single representative $\mathbf{x}_i$. To satisfy \textbf{Conservation of Attention Mass}, we need to introduce a bias $B$:
\begin{equation}
    \exp(\mathbf{q}^\top \mathbf{x}_i + B) \approx \sum_{j \in \Omega_i} \exp(\mathbf{q}^\top \mathbf{x}_j) \approx f_i \cdot \exp(\mathbf{q}^\top \mathbf{x}_i)
\end{equation}
This leads to the correction term: $B = \log(f_i)$.

\textbf{Conclusion:} $\log(f_i)$ is a mathematically necessary correction term. It ensures that compressed background regions maintain correct semantic weight during attention computation, fundamentally suppressing hallucinations caused by missing background information (i.e., \textbf{Attention Dilution}).

\subsection{Adaptive Quantization: A Fixed-Distortion Variable-Rate Perspective}
\label{app:adaptive_quantification}

We reconstruct the Token Compression problem as an optimal quantization problem. Unlike existing methods (e.g., StreamMem) that enforce a \textbf{Fixed-Rate} constraint (i.e., retaining a fixed $K$ tokens per frame), we adopt a \textbf{Fixed-Distortion} constraint:
\begin{equation}
    \min_{\mathbf{Z}_t} |\mathbf{Z}_t| \quad \text{s.t.} \quad \|\mathbf{X}_t - \text{Reconstruct}(\mathbf{Z}_t)\|_\infty \le \epsilon
\end{equation}
Under this framework, the number of retained tokens $M_t$ naturally becomes an estimator of the frame's \textbf{intrinsic information entropy}: $M_t(\mathbf{X}_t) \propto \mathcal{H}(\mathbf{X}_t)$. This allows the algorithm to adaptively allocate the KV budget based on the richness of video content (e.g., static vs. dynamic scenes), conceptually aligning with the objectives of \textbf{Rate-Distortion Theory} rather than claiming strict theoretical optimality.

\subsection{Algorithmic Implementation Details}
\label{app:implementation_details}

This section details the algorithmic flow of the proposed two-stage compression framework.

\paragraph{Phase 1: Inter-Frame Filtering.}
Aims to rapidly eliminate repetitive information in the temporal dimension.
\begin{itemize}
    \item \textbf{Input:} Current frame token features $\mathbf{V}^{(t)} = \{\mathbf{v}_i^{(t)}\}$, previous frame features $\mathbf{V}^{(t-1)}$.
    \item \textbf{Similarity Metric:} Compute cosine similarity at corresponding positions:
    \begin{equation}
        \text{sim}(\mathbf{v}_i^{(t)}, \mathbf{v}_i^{(t-1)}) = \frac{\mathbf{v}_i^{(t)} \cdot \mathbf{v}_i^{(t-1)}}{\|\mathbf{v}_i^{(t)}\| \|\mathbf{v}_i^{(t-1)}\|}
    \end{equation}
    \item \textbf{Decision:} If $\text{sim} > \theta_{\text{inter}}$, it is deemed redundant and discarded; otherwise, it is retained for the next phase.
\end{itemize}

\paragraph{Phase 2: Intra-Frame Aggregation.}
Spatially aggregates surviving tokens via matching and weighted reconstruction.

\textbf{Step 1: Bipartite Matching Decision.}
We employ an iterative bipartite matching strategy \cite{bolya2022token} to determine merge sets.
Partition current frame tokens into sets $\mathcal{A}$ (source) and $\mathcal{B}$ (target). Compute the similarity matrix between sets; if $\text{sim}(\mathbf{a}, \mathbf{b}) > \theta_{\text{intra}}$ and they are mutual best matches, establish a merge edge. This process can be iterated to increase compression rate.

\textbf{Step 2: Gated Similarity Attention (GSim).}
To smoothly fuse information and compute the fusion count $f_i$, we employ a gating mechanism. Let the surviving token set be $\mathbf{X}_{S}$ and the raw token set be $\mathbf{X}_{raw}$.
\begin{enumerate}[nosep]
    \item \textbf{Gated Similarity Calculation:}
    \begin{equation}
    \mathbf{G}_{ij} = \begin{cases} \mathbf{x}_i^T \mathbf{z}_j & \text{if } \mathbf{x}_i^T \mathbf{z}_j \geq \kappa \\ -\infty & \text{otherwise} \end{cases}, \quad \text{where } \mathbf{x}_i \in \mathbf{X}_{S}, \mathbf{z}_j \in \mathbf{X}_{raw}
    \end{equation}

    \item \textbf{Normalized Weight Assignment:}
    Use temperature $\tau$ to compute the soft assignment matrix $\mathbf{M}$:
    \begin{equation}
    \mathbf{M}_{ij} = \frac{\exp(\mathbf{G}_{ij} / \tau)}{\sum_{k} \exp(\mathbf{G}_{kj} / \tau)}
    \end{equation}
    Here, $\mathbf{M}_{ij}$ represents the contribution of raw token $\mathbf{z}_j$ to survivor token $\mathbf{x}_i$.
    
    \item \textbf{Fusion Count \& Feature Reconstruction:}
    Compute the effective fusion number $f_i$ for each survivor (for subsequent Logit Bias):
    \begin{equation}
    f_i = \sum_{j} \mathbf{M}_{ij}
    \end{equation}
    The final feature $\mathbf{y}_i$ is a weighted combination of raw features:
    \begin{equation}
    \mathbf{y}_i = \frac{\sum_{j} \mathbf{M}_{ij} \mathbf{z}_j}{f_i + \epsilon}
    \end{equation}

\end{enumerate}

Through these steps, we not only compress feature dimensions but, crucially, preserve the count information $f_i$ of tokens, providing the necessary parameters for \textbf{Attention Conservation} in subsequent attention layers.

\section{Managing Structural Sensitivity: Rigorous Derivation of Budget Allocation Strategy}
\label{app:budget_theory}

This section provides the rigorous mathematical derivation for the proposed \textbf{Sensitivity-Aware Budget Allocation} strategy. We formalize KV Cache compression as a constrained minimization problem of layer-wise attention output error. We first derive a \textbf{Linear Upper Bound} of this error based on matrix perturbation analysis, then prove from a statistical perspective that a \textbf{Variance-Stabilized Logarithmic Surrogate} is required to achieve robust optimization under heavy-tailed LLM activation distributions.

\subsection{Problem Formulation}

Consider the Multi-Head Attention (MHA) module of the $l$-th layer. Let $\mathcal{O}^{(l)} \in \mathbb{R}^{T \times d}$ be the final output projection of this layer. Its computation is a linear combination of outputs from each attention head:
\begin{equation}
    \mathcal{O}^{(l)} = \left[ \text{Concat}_{h=1}^H \left( \mathbf{A}^{(l,h)} \mathbf{V}^{(l,h)} \right) \right] \mathbf{W}_{\mathcal{O}}^{(l)}
\end{equation}
where $\mathbf{A}^{(l,h)}$ is the normalized attention probability matrix, $\mathbf{V}^{(l,h)}$ is the value vector, and $\mathbf{W}_{\mathcal{O}}^{(l)}$ is the output projection matrix.
When performing KV Cache pruning, we introduce a binary mask tensor $\mathbf{M} \in \{0,1\}^{L \times H \times T}$. The pruned attention matrix $\hat{\mathbf{A}}$ undergoes Softmax renormalization, producing the approximate output $\hat{\mathcal{O}}^{(l)}$.

Our optimization goal is to minimize the $L_1$ norm of the cumulative approximation error under the global memory budget $B_{\text{total}}$:
\begin{equation}
    \min_{\mathbf{M}} \sum_{l=1}^L \left\| \mathcal{O}^{(l)} - \hat{\mathcal{O}}^{(l)} \right\|_1 \quad \text{s.t.} \quad \sum_{l=1}^L \text{NNZ}(\mathbf{M}^{(l)}) \le B_{\text{total}}
\end{equation}
where $\text{NNZ}(\cdot)$ denotes the number of non-zero elements.

\subsection{Derivation of the Error Upper Bound}

Based on the compatibility of matrix norms and fundamental lemmas from AdaKV \cite{feng2024ada}, we derive the upper bound for single-layer error.
Let the error term be $\mathcal{E}^{(l)} = \|\mathcal{O}^{(l)} - \hat{\mathcal{O}}^{(l)}\|_1$. Using Hölder's inequality, the error can be separated into structural parameters and activations:
\begin{equation}
    \mathcal{E}^{(l)} \le \underbrace{\left\| (\mathbf{W}_{\mathcal{O}}^{(l)})^\top \right\|_\infty}_{C_1} \cdot \sum_{h=1}^H \left\| (\mathbf{A}^{(l,h)} - \hat{\mathbf{A}}^{(l,h)}) \mathbf{V}^{(l,h)} \right\|_1
\end{equation}
where $C_1$ is a structural constant related only to model weights.

The key lies in bounding the perturbation of attention weights $\|\mathbf{A} - \hat{\mathbf{A}}\|_1$. Let $\epsilon_h = \sum_{i \in \mathcal{S}_{\text{pruned}}} p_{h,i}$ be the sum of original probabilities of pruned tokens. Considering the renormalization effect of Softmax (i.e., retained weights are scaled by $1/(1-\epsilon_h)$), according to Theorem 1 in \cite{shen2025lava}, we obtain a strict linear upper bound:
\begin{equation}
\label{eq:linear_bound}
    \mathcal{E}^{(l)} \le 2 C_1 \sum_{h=1}^H \sum_{i \in \mathcal{S}_{\text{pruned}}} \left( p_{h,i} \cdot \|\mathbf{v}_{h,i}\|_1 \right)
\end{equation}
This formula reveals the theoretical optimal greedy strategy: one should prioritize retaining tokens where the \textbf{product of attention weight $p$ and value vector norm $\|\mathbf{v}\|$ is maximized}.

\subsection{Robust Logarithmic Surrogate}

Although Eq. (\ref{eq:linear_bound}) provides a theoretical optimum, directly using the linear product $p \cdot \|\mathbf{v}\|$ as a scoring function is numerically unstable. This is attributed to the \textbf{Distributional Heterogeneity} of the MLLM feature space.

\paragraph{Statistical Instability.}
LLM activations typically follow a significant heavy-tailed distribution. Outlier layers may have $\|\mathbf{v}\|$ magnitudes orders of magnitude higher than the average. If budgets are allocated directly based on the linear upper bound, these outlier layers will trigger a Winner-takes-all effect, hijacking the vast majority of the budget and causing layer collapse in semantically critical but numerically smaller layers. Furthermore, model performance error sensitivity typically exhibits non-linear saturation.

To address this, we transform the optimization objective from minimizing linear error upper bound to maximizing logarithmic utility. Specifically, we introduce a \textbf{Variance-Stabilizing Transformation} as an empirical heuristic. The choice of the logarithmic function ensures diminishing returns for overly dense regions, thereby yielding a robust Pareto frontier. We construct the dynamic surrogate metric $\mathcal{M}_{\text{dyn}}$ as follows:

\paragraph{Magnitude Surrogate: $\log \|\mathbf{V}\|$}
Apply a logarithmic transformation to feature intensity to smooth inter-layer magnitude differences:
\begin{equation}
    \text{Score}_{\text{mag}}^{(l)} = \log(\|\mathbf{V}^{(l)}\|_1 + \delta)
\end{equation}
This is statistically equivalent to stabilization operations for heteroscedastic data, ensuring budget allocation is robust to outliers.

\paragraph{Complexity Surrogate: $\log(1+\mathcal{H})$}
The cumulative probability error in Eq. (\ref{eq:linear_bound}) is negatively correlated with attention distribution sparsity. Shannon entropy $\mathcal{H}$ is a global measure of sparsity.
\begin{itemize}
    \item \textbf{Necessity:} High-entropy layers (flat distribution) imply information diffusion, where probability loss $\epsilon$ from pruning decreases slowly, requiring more budget coverage.
    \item \textbf{Diminishing Returns:} However, attempting to fully fit a long-tail distribution is prohibitively expensive with limited returns.
\end{itemize}
Using $\log(1+\mathcal{H})$ as a complexity penalty factor acknowledges the importance of high-entropy layers while suppressing excessive investment in overly uniform distributions (often corresponding to invalid background noise) via logarithmic function concavity.

In summary, we construct the layer importance score $\mathcal{M}^{(l)}_{\text{dyn}}$ as the coupling of two robust factors:
\begin{equation}
    \mathcal{M}^{(l)}_{\text{dyn}} \propto \underbrace{\log(1+\mathcal{H}(\mathbf{A}^{(l)}))}_{\text{Complexity (Dispersion)}} \cdot \underbrace{\log(\|\mathbf{V}^{(l)}\|_1)}_{\text{Magnitude (Intensity)}}
\end{equation}

\subsection{Derivation of Structural Sensitivity Prior}
\label{app:structural_sensitivity}

Existing KV Cache compression methods (e.g., H2O, SnapKV) primarily focus on local approximation error $\|\mathbf{A}^{(l)} - \hat{\mathbf{A}}^{(l)}\|$, implicitly assuming that every layer contributes equally to the final output. However, the deep non-linear structure of Transformers dictates that perturbations at different levels have significantly heterogeneous impacts on the final prediction. Here, we use a first-order Taylor expansion to derive \textbf{Structural Sensitivity}, arguing for its necessity as a static prior.

\subsubsection{Decoupling Global Error}
Let the model's final loss function be $\mathcal{L}$ (or output  $\mathcal{O}$). The local perturbation introduced by KV Cache pruning at layer $l$ is $\Delta \mathbf{A}^{(l)}$. According to the chain rule of multivariate calculus, the total error $\Delta \mathcal{L}$ can be approximated as the weighted sum of layer-wise perturbations:
\begin{equation}
    \Delta \mathcal{L} \approx \sum_{l=1}^L \text{tr}\left( \left(\frac{\partial \mathcal{L}}{\partial \mathbf{A}^{(l)}}\right)^\top \Delta \mathbf{A}^{(l)} \right)
\end{equation}
Applying the Cauchy-Schwarz Inequality, we obtain the error upper bound:
\begin{equation}
    |\Delta \mathcal{L}| \le \sum_{l=1}^L \underbrace{\left\| \frac{\partial \mathcal{L}}{\partial \mathbf{A}^{(l)}} \right\|_F}_{\text{Global Sensitivity } S_l} \cdot \underbrace{\left\| \Delta \mathbf{A}^{(l)} \right\|_F}_{\text{Local Distortion } \mathcal{E}^{(l)}}
\end{equation}
This inequality reveals the dual dependency of optimal budget allocation: budgets should be allocated not only to layers with \textbf{large local error} (high $\mathcal{E}^{(l)}$, corresponding to high Entropy), but also to layers with \textbf{large impact on global output} (high $S_l$).

\subsubsection{Gradient Surrogate and Structural Invariance Assumption}
Online calculation of the Jacobian norm $\left\| \frac{\partial \mathcal{L}}{\partial \mathbf{A}^{(l)}} \right\|_F$ requires full backpropagation, which is unacceptable during inference. Therefore, we introduce an \textbf{Assumption}:

\textit{
Within a specific task domain, the relative importance of Transformer layers to the output is primarily determined by the model architecture and fluctuates minimally with input samples.}

Based on this, we approximate sensitivity $S_l$ as the expected gradient norm on a calibration dataset $\mathcal{D}_{\text{cal}}$:
\begin{equation}
    S_l \approx \mathbb{E}_{\mathbf{x} \sim \mathcal{D}_{\text{cal}}} \left[ \left\| \nabla_{\mathbf{A}^{(l)}} F(\mathbf{x}) \right\|_F \right]
\end{equation}
To avoid data leakage, although our algorithm is developed for video, we select the single-image dataset LLaVA-Pretrain \cite{liu2023visual} as the calibration set.
This offline-computed $S_l$ captures the topological fingerprint of the model. For instance, experiments observe that shallow layers typically exhibit larger gradient norms (large $S_l$), implying that minor pruning errors in shallow layers are significantly amplified during deep propagation (Error Avalanche Effect). This provides direct theoretical proof for introducing the static weight $\lambda_s S^{(l)}$ in our budget allocation formula: \textbf{To minimize the global upper bound $|\Delta \mathcal{L}|$, budget $b_l$ must be positively correlated with sensitivity $S_l$.}

\subsection{Bayesian Interpretation of the Hybrid Strategy}
Finally, we interpret the proposed hybrid allocation strategy as a \textbf{Maximum A Posteriori (MAP) estimation in the log-domain}. We view the budget proportion $\rho^{(l)}$ of each layer as posterior importance, determined by three terms:

\begin{equation}
    \underbrace{\rho^{(l)}}_{\text{Log-Posterior}} \propto \underbrace{\mathcal{M}_{\text{dyn}}^{(l)}}_{\text{Log-Likelihood}} + \lambda_s \underbrace{\mathcal{S}^{(l)}}_{\text{Log-Prior}} + \lambda_u \underbrace{\mathcal{U}}_{\text{Regularization}}
\end{equation}

\begin{enumerate}
    \item \textbf{Data Likelihood $\mathcal{M}_{\text{dyn}}^{(l)}$:} 
    The robust error surrogate derived above. It reflects data-driven dynamic evidence: how much incompressible effective information is contained in the layer given the current context.

    \item \textbf{Structural Prior $\mathcal{S}^{(l)}$:} 
    The offline-computed gradient sensitivity $S_l$. It encodes the inherent topological preferences of the model architecture, correcting potential biases from relying solely on online statistics.
    
    \item \textbf{MaxEnt Regularization $\mathcal{U}$:} 
    A uniform distribution term. It corresponds to the Maximum Entropy Prior (Uninformative Prior) in the Bayesian framework, acting as a smoothing factor to prevent budget allocation degeneration due to the failure of any single metric, ensuring the lower-bound stability of the strategy.

\end{enumerate}

In this Bayesian formulation, the linear trade-off hyperparameters ($\lambda_s, \lambda_u$) effectively correspond to the \textbf{precision (inverse variance)} of the respective log-domain priors. This interpretation allows us to explicitly control the confidence placed on static architecture features versus online dynamic observations.

\section{Supplementary Experimental Details and Hyperparameter Configurations}
\label{app:Hyperparameter}
\subsection{Comparison with Various Compression Methods at Different Ratios}
\label{app:mlvu_details}
In the evaluation of Qwen2-VL-7B~\cite{wang2024qwen2}, we compare our proposed method with a comprehensive set of compression baselines, categorized as follows:
\begin{enumerate}
    \item \textbf{Visual Token Compression:} e.g., FastV~\cite{chen2024image}, which prunes tokens based on attention scores.
    \item \textbf{LLM-Oriented KV Compression:} Algorithms originally designed for Large Language Models, including SnapKV~\cite{li2024snapkv} and the standard InfiniPot~\cite{kim2024infinipot}.
    \item \textbf{MLLM-Specific KV Compression:} State-of-the-art methods customized for Multimodal LLMs, specifically InfiniPot-V~\cite{kim2025infinipot} and StreamMem~\cite{yang2025streammem}.
    \item \textbf{Uniform Sampling:} A baseline method that retains KV pairs uniformly.
\end{enumerate}
Following the evaluation protocols of InfiniPot-V~\cite{kim2025infinipot} and StreamMem~\cite{yang2025streammem}, we report results on the official validation set (\texttt{dev}) of the MLVU benchmark. Given that our method involves layer-wise dynamic budget allocation, to ensure fairness and comparability, we strictly control the global total capacity of the KV Cache to be equivalent to Average Budget per Layer levels of 3,000, 6,000, 12,000, and 24,000.

\subsubsection{Experimental Parameter Settings}
\label{sec:hyperparams}

For the budget allocation strategy, the default weight coefficients are set as $(\lambda_u, \lambda_s, \lambda_e) = (0.8, 0.166, 0.034)$. 
To maximize the synergy of our two-stage compression framework (Token Reduction + KV Pruning), we adaptively adjust the inter-frame ($\tau_{\text{inter}}$) and intra-frame ($\tau_{\text{intra}}$) similarity thresholds based on the specific memory budget. 
The detailed configurations for various budget levels (3K, 6K, 12K, and 24K) are summarized in Table~\ref{tab:hyper_settings}.

\begin{table}[h]
    \centering
    \caption{\textbf{Hyperparameter settings for different KV budgets.} To achieve optimal performance under stricter budgets, we adopt slightly more aggressive filtering thresholds.}
    \label{tab:hyper_settings}
    \vspace{0.1in}
    \begin{tabular}{c c c}
    \toprule
    \textbf{KV Budget} & \textbf{Inter-Frame} ($\tau_{\text{inter}}$) & \textbf{Intra-Frame} ($\tau_{\text{intra}}$) \\
    \midrule
    3000  (3K) & 0.15 & 0.45 \\
    6000  (6K) & 0.20 & 0.55 \\
    12000 (12K)& 0.25 & 0.55 \\
    24000 (24K)& 0.30 & 0.60 \\
    \bottomrule
    \end{tabular}
\end{table}

\subsubsection{Experimental Results}
Detailed performance comparisons on the MLVU benchmark are presented in Table~\ref{tab:mlvu_full_results}.
As shown, our method consistently outperforms all baselines across all budget levels. Notably, at 12K and 24K budgets, our method achieves accuracies of \textbf{68.69\%} and \textbf{69.43\%}, respectively, surpassing even the \textbf{Full KV} baseline (65.85\%). This demonstrates that our structure-aware filtering mechanism effectively removes noise while preserving semantically dense states.

\subsection{Detailed Experimental Data and Analysis of Compression Strategy Trade-offs}
\label{app:compression_tradeoff}

In this section, we provide detailed experimental settings and numerical data for the compression strategy performance comparison graph (Figure \ref{compression_performance}), and explain our adopted compression method from an information-theoretic view.

\subsubsection{Experimental Setup and Hyperparameters}
Experiments were conducted using the \textbf{Qwen2.5-VL-3B} \cite{bai2025qwen2} model on the \textbf{VideoMME} \cite{fu2025video} long video benchmark. We compared three strategies:
\begin{enumerate}
    \item \textbf{KV Cache Compression Only:} Prunes the KV cache using only sparsity metrics (Attention Sparsity).
    \item \textbf{Two-Stage Compression (Hybrid Strategy):} Reduces input tokens via Inter-Frame Filtering and Intra-Frame Aggregation. We set the inter-frame similarity threshold $\tau_{\text{inter}} = 0.25$ and the default intra-frame merge threshold $\tau_{\text{intra}} = 0.75$. The only exception is under extremely low budgets (KV cache size = 500), where we relax $\tau_{\text{intra}}$ to $0.70$ to accommodate strict capacity limits, followed by streaming prefill and KV cache compression if the budget limit is exceeded.
    \item \textbf{Token Compression Only:} Uses the same token compression strategy as the \textbf{Hybrid Strategy}. Subsequently, if limits are exceeded, the remaining tokens for each frame are proportionally compressed to fit exactly into a window of the same size as the KV cache using our intra-frame token compression algorithm.
\end{enumerate}

\subsubsection{Detailed Performance Data}
Table \ref{tab:compression_data} presents specific accuracy data under different KV cache budgets. For comparison, we use InfiniPot-V \cite{kim2025infinipot}, StreamMem \cite{yang2025streammem}, and the original full model (Original Model, Acc=50.56\%) as baselines.

\subsubsection{Results Analysis}
Analyzing Table \ref{tab:compression_data} and the trend graph leads to the following key observations:

\paragraph{Saturation Effect of KV Cache Compression:}
The method relying solely on KV cache compression (Orange line) is robust at low budgets, but performance growth stagnates as the budget increases ($>3000$), stabilizing near the level of the original model (Original Model, 50.56\%). This indicates that while retaining critical KV pairs via sparsity maintains performance, it struggles to break the context processing bottleneck of the original model.

\paragraph{Performance Gains from Token Compression:}
In contrast, the Token Compression strategy (Blue line) exhibits superior performance at higher budgets, significantly surpassing the original model (e.g., reaching 52.89\% at a budget of 12000). We attribute this to two factors:
\begin{itemize}
    \item \textbf{Redundancy Removal and Dynamic Capture:} The inter-frame filtering mechanism effectively removes static redundancy, allowing the model to focus on dynamic changes in the video.
    \item \textbf{Attention Concentration:} Through intra-frame merging and subsequent attention score correction, we shorten the effective context length. This avoids the Attention Dilution problem common in long sequences, enabling the model to capture key cues more sharply, thus achieving accuracy superior to full input in long video understanding tasks.
\end{itemize}

Ultimately, our Two-Stage Hybrid method (Green line) combines the robustness of KV cache compression with the high information density of token compression, achieving an excellent performance trade-off across the entire budget range.

\subsubsection{An Information-Theoretic Perspective}

From an information-theoretic perspective, our hybrid framework can be interpreted as an effective \textbf{Joint Source-Channel Coding} scheme:
\begin{itemize}
    \item \textbf{Token Compression (Source Coding):} Exploits data autocorrelation (similarity) to perform lossless (or low-distortion) information aggregation via feature reconstruction and logit correction, maximizing the information entropy density of input tokens.
    
    \item \textbf{KV Cache Compression (Channel Allocation):} Exploits attention sparsity to dynamically allocate limited storage channel capacity (KV cache budget) to features with high signal-to-noise ratios (high-sensitivity layers, high-activation tokens). Unlike the explicit fusion in Token Compression, this is a \textbf{Selective Memory} mechanism based on importance evaluation.
\end{itemize}

The experimental results in Figure \ref{compression_performance} corroborate this theory: token compression alone preserves more complete information at high budgets (Blue curve), while KV cache compression alone is more robust at screening key information at low budgets (Orange curve). Our hybrid method (Green curve) combines the advantages of both, achieving a superior performance-memory trade-off across the full budget range.
\subsection{Empirical Analysis of Layer-wise Statistics}
\label{app:stat_analysis}

To provide empirical insights into the proposed hybrid budget allocation strategy, we conducted an in-depth statistical analysis of the layer-wise behavior of Qwen2.5-VL-3B \cite{bai2025qwen2}. This section details the data sources, collection methods, and statistical regularities underlying Figure \ref{fig:stats}, justifying the necessity of integrating structural priors with dynamic context metrics.

\begin{table*}[h]
    \centering
    \caption{\textbf{Performance comparison on MLVU (Qwen2-VL-7B) across different KV budgets.} We report the average accuracy (\%). Full KV represents the performance of the original model without compression. \textbf{Bold} indicates the best performance under compression.}
    \label{tab:mlvu_full_results}
    \vspace{0.1in}
    \resizebox{0.6\linewidth}{!}{
    \begin{tabular}{lccccc}
    \toprule
    \multirow{2}{*}{\textbf{Method}} & \multicolumn{4}{c}{\textbf{KV Budget}} & \multirow{2}{*}{\textbf{Avg.}} \\
    
    \cmidrule(lr){2-5}
    
     & \textbf{3K} & \textbf{6K} & \textbf{12K} & \textbf{24K} & \\
    \midrule
    
    \multicolumn{6}{l}{\textit{Baselines}} \\
    Uniform & 55.59 & 57.00 & 60.36 & 64.18 & 59.28 \\
    FastV~\cite{chen2024image} & 55.01 & 57.60 & - & - & - \\
    SnapKV~\cite{li2024snapkv} & 58.99 & 60.66 & 61.31 & 64.37 & 61.33 \\
    InfiniPot~\cite{kim2024infinipot} & 58.36 & 59.11 & 60.70 & 63.71 & 60.47 \\
    \midrule
    
    \multicolumn{6}{l}{\textit{State-of-the-Art MLLM Methods}} \\
    InfiniPot-V~\cite{kim2025infinipot} & 64.70 & 65.82 & 65.99 & 65.73 & 65.56 \\
    StreamMem~\cite{yang2025streammem} & - & 65.90 & 66.00 & 66.30 & - \\
    \midrule
    
    \rowcolor{gray!10} \textbf{Video-MemComp} (Ours) & \textbf{65.48} & \textbf{67.10} & \textbf{68.69} & \textbf{69.43} & \textbf{67.53} \\
    \midrule
    
    \textit{Full KV (Reference)} & \multicolumn{4}{c}{65.85} & 65.85 \\
    \bottomrule
    \end{tabular}
    }
\end{table*}

\subsubsection{Data Collection Protocols}
To balance \textbf{Generalizability} and \textbf{Domain Adaptability}, we collected static and dynamic statistics on two independent data distributions:

\begin{itemize}
    \item \textbf{Static Structural Sensitivity ($S_l$):} 
    To capture the inherent topological importance of the model architecture and avoid overfitting to specific video domains, we used the single-image pre-training dataset LLaVA-Pretrain \cite{liu2023visual}. We randomly sampled 100 examples and computed the expected gradient norm of the attention output $\mathbf{A}^{(l)}$ with respect to the final loss function, $\mathbb{E}[\|\frac{\partial \mathcal{L}}{\partial \mathbf{A}^{(l)}}\|_F]$, via backpropagation as the static sensitivity prior.

    \item \textbf{Dynamic Context Metrics ($H$ and $\|V\|$):} 
    To reflect actual activation patterns when processing long-form streaming data, we used the long video understanding benchmark VideoMME \cite{fu2025video}. During inference, we tracked 10 randomly sampled video streams, recording the real-time attention distribution entropy (Entropy) and the feature magnitude (Feature Norm) of value vectors for each layer.

\end{itemize}
\begin{table*}[h]
\centering
\caption{\textbf{Comparison of Compression Strategy Performance on VideoMME.} We report model accuracy (\%) under different KV budgets. The baseline performance of the original model (Qwen2.5-VL-3B without compression) is 50.56\%.}
\label{tab:compression_data}
\resizebox{0.95\textwidth}{!}{%
\begin{tabular}{c|ccc|cc}
\toprule
\textbf{KV Budget} & \textbf{Token Comp.} & \textbf{KV Cache Comp.} & \textbf{\textbf{Video-MemComp} (Ours) (Two-Stage)} & \textbf{InfiniPot-V} & \textbf{StreamMem} \\
\midrule
500   & 45.56 & 45.89 & \textbf{47.78} & 47.00 & 44.00 \\
3000  & 49.33 & 49.44 & \textbf{50.33} & 48.00 & 44.60 \\
6000  & 51.22 & 50.78 & \textbf{53.22} & 48.00 & 49.10 \\
12000 & 52.89 & 51.11 & \textbf{53.44} & 49.67 & 50.00 \\
\bottomrule
\end{tabular}%
}
\end{table*}

\subsubsection{Static Prior Analysis: Gradient Norm Distribution}
Table \ref{tab:grad_stats} details the statistical characteristics of gradient norms for each layer. We observe two key phenomena:
\begin{enumerate}
    \item \textbf{Gradient Decay Trend:} Shallow layers (Layers 0-10) exhibit extremely high mean gradients. This confirms the Error Avalanche Effect hypothesis—minor perturbations in shallow layers are amplified layer-by-layer during deep propagation. Therefore, budget allocation must be skewed towards shallow layers.
    \item \textbf{Inter-Sample Variance:} Notably, the Relative Standard Deviation (RSD) of shallow layers is generally high ($>30\%$). This implies that while shallow layers are important overall, the reliance of different samples on specific layers fluctuates significantly. This strongly proves that relying solely on static priors is insufficient; online dynamic metrics must be introduced to adapt to specific sample variations.
\end{enumerate}

\begin{table*}[h]
\centering
\caption{\textbf{Distribution of Dynamic Metrics for Qwen2.5-VL-3B during Online Inference.} Data is sampled from VideoMME streams. $H$ denotes the Shannon entropy of the attention probability distribution, and $\log \|V\|$ is the log-norm of the value vectors. \textbf{Layers 3-28} exhibit extremely low entropy values, constituting the primary compression space.}
\label{tab:dynamic_stats}
\resizebox{0.7\textwidth}{!}{%
\begin{tabular}{ccc|ccc}
\toprule
\textbf{Layer} & \textbf{Entropy ($H$)} & \textbf{Log Feature Norm} & \textbf{Layer} & \textbf{Entropy ($H$)} & \textbf{Log Feature Norm} \\
\midrule
0 & 0.701 & 1.44 & 18 & $4.08 \times 10^{-27}$ & 2.14 \\
1 & 1.104 & 0.97 & 19 & $1.48 \times 10^{-24}$ & 2.09 \\
2 & 0.766 & 1.04 & 20 & $5.45 \times 10^{-23}$ & 2.17 \\
3 & $2.82 \times 10^{-31}$ & 1.80 & 21 & $1.32 \times 10^{-21}$ & 2.19 \\
4 & $1.49 \times 10^{-35}$ & 1.65 & 22 & $1.07 \times 10^{-20}$ & 2.19 \\
5 & $3.94 \times 10^{-36}$ & 1.73 & 23 & $3.62 \times 10^{-28}$ & 2.09 \\
6 & $1.81 \times 10^{-29}$ & 1.95 & 24 & $2.29 \times 10^{-27}$ & 2.03 \\
7 & $6.49 \times 10^{-26}$ & 2.00 & 25 & $1.80 \times 10^{-20}$ & 2.14 \\
8 & $6.99 \times 10^{-23}$ & 2.03 & 26 & $2.04 \times 10^{-25}$ & 2.30 \\
9 & $1.21 \times 10^{-22}$ & 1.90 & 27 & $3.95 \times 10^{-16}$ & 2.36 \\
10 & $3.28 \times 10^{-23}$ & 2.09 & 28 & $1.69 \times 10^{-22}$ & 2.36 \\
11 & $6.52 \times 10^{-23}$ & 2.00 & 29 & $1.44 \times 10^{-21}$ & 2.20 \\
12 & $1.62 \times 10^{-23}$ & 2.05 & 30 & $6.10 \times 10^{-18}$ & 2.50 \\
13 & $7.88 \times 10^{-28}$ & 1.95 & 31 & 0.026 & 2.64 \\
14 & $1.88 \times 10^{-26}$ & 1.92 & 32 & 0.003 & 2.92 \\
15 & $2.20 \times 10^{-31}$ & 1.99 & 33 & 0.001 & 3.20 \\
16 & $1.33 \times 10^{-23}$ & 1.98 & 34 & 0.005 & 2.66 \\
17 & $1.14 \times 10^{-23}$ & 2.09 & 35 & $3.23 \times 10^{-5}$ & 2.59 \\
\bottomrule
\end{tabular}%
}
\end{table*}

\subsubsection{Dynamic Behavior Analysis: Three-Stage Evolution}

Table \ref{tab:dynamic_stats} reveals a distinct three-stage feature distribution during online inference, which aligns well with the intuition behind our hybrid allocation strategy:

\begin{enumerate}
    \item \textbf{Feature Extraction Phase (Layers 0-2):} 
    This phase exhibits higher entropy and extremely large gradients, corresponding to the initial encoding of local visual features. Our strategy ensures an ample budget here via high gradient priors ($S_l$).
    
    \item \textbf{Sparsity Redundancy Zone (Layers 3-30):} 
    This is the most striking finding. In the vast middle layers, attention entropy drops sharply to extremely low levels (on the order of $10^{-20}$). This indicates that the attention mechanism is highly concentrated on a very small number of tokens, revealing significant sparsity redundancy. Our dynamic entropy metric ($\log(1+\mathcal{H})$) effectively identifies this zone and performs aggressive pruning, thereby saving substantial budget.
    
    \item \textbf{Semantic Aggregation Phase (Layers 31-35):} 
    In deeper layers, the entropy slightly increases, while the feature norm reaches its peak. However, the gradient sensitivity $\mathcal{S}_l$ remains relatively low. Although the feature magnitude is high, the combination of low gradient sensitivity and still relatively low entropy results in a limited budget allocation for these layers.
\end{enumerate}

\subsection{Model Adaptation and Hyperparameter Configuration}
\label{app:hyperparams}

Given significant differences in video duration distribution, task complexity, and modality dependency across benchmarks, we implemented targeted configuration optimizations for each Model-Benchmark combination. For example, LLaVA-OV employed YaRN \cite{pengyarn} length extrapolation for handling extremely long videos, dynamically adjusting compression thresholds based on duration; InternVL adjusted frame sampling strategies according to its input resolution characteristics. Specific hyperparameter configurations are detailed in Table \ref{tab:hyperparams}.
\begin{table*}[t]
\centering
\caption{\textbf{Statistical Analysis of Structural Sensitivity ($S_l$) for Qwen2.5-VL-3B \cite{bai2025qwen2}.} Data is calculated based on LLaVA-Pretrain \cite{liu2023visual}. A higher Mean value indicates the layer is more sensitive to pruning errors. RSD (Relative Standard Deviation) reflects the fluctuation of importance across different samples.}
\label{tab:grad_stats}
\resizebox{0.7\textwidth}{!}{%
\begin{tabular}{cccc|cccc}
\toprule
\textbf{Layer} & \textbf{Mean ($S_l$)} & \textbf{Std} & \textbf{RSD (\%)} & \textbf{Layer} & \textbf{Mean ($S_l$)} & \textbf{Std} & \textbf{RSD (\%)} \\
\midrule
0 & 2.2740e-04 & 7.0700e-05 & 31.09 & 18 & 1.2093e-04 & 2.7035e-05 & 22.36 \\
1 & 1.6284e-04 & 5.2962e-05 & 32.52 & 19 & 1.1845e-04 & 2.5720e-05 & 21.71 \\
2 & 1.5621e-04 & 5.1101e-05 & 32.71 & 20 & 1.1420e-04 & 2.4370e-05 & 21.34 \\
3 & 1.5206e-04 & 4.9749e-05 & 32.72 & 21 & 1.1110e-04 & 2.2826e-05 & 20.54 \\
4 & 1.4615e-04 & 4.7366e-05 & 32.41 & 22 & 1.0948e-04 & 2.1954e-05 & 20.05 \\
5 & 1.4100e-04 & 4.4595e-05 & 31.63 & 23 & 1.0638e-04 & 2.0570e-05 & 19.34 \\
6 & 1.3461e-04 & 4.2124e-05 & 31.29 & 24 & 9.9707e-05 & 1.8553e-05 & 18.61 \\
7 & 1.3132e-04 & 4.0984e-05 & 31.21 & 25 & 9.5272e-05 & 1.7394e-05 & 18.26 \\
8 & 1.2841e-04 & 3.9545e-05 & 30.80 & 26 & 8.8787e-05 & 1.5637e-05 & 17.61 \\
9 & 1.2703e-04 & 3.8906e-05 & 30.63 & 27 & 8.2707e-05 & 1.4318e-05 & 17.31 \\
10 & 1.2250e-04 & 3.5122e-05 & 28.67 & 28 & 7.6103e-05 & 1.2959e-05 & 17.03 \\
11 & 1.1930e-04 & 3.0953e-05 & 25.94 & 29 & 6.8331e-05 & 1.1477e-05 & 16.80 \\
12 & 1.1377e-04 & 2.8229e-05 & 24.81 & 30 & 5.8675e-05 & 9.3638e-06 & 15.96 \\
13 & 1.2050e-04 & 2.9743e-05 & 24.68 & 31 & 4.9806e-05 & 7.7510e-06 & 15.56 \\
14 & 1.1802e-04 & 2.8380e-05 & 24.05 & 32 & 4.1294e-05 & 6.4592e-06 & 15.64 \\
15 & 1.2393e-04 & 3.0301e-05 & 24.45 & 33 & 3.4630e-05 & 5.5330e-06 & 15.98 \\
16 & 1.2283e-04 & 2.9177e-05 & 23.75 & 34 & 2.8634e-05 & 4.5449e-06 & 15.87 \\
17 & 1.2326e-04 & 2.8743e-05 & 23.32 & 35 & 2.4843e-05 & 3.9889e-06 & 16.06 \\
\bottomrule
\end{tabular}%
}
\end{table*}
\begin{table*}[h]
\centering
\caption{\textbf{Detailed hyperparameter configurations for different models across benchmarks.} $\tau_{\text{inter}}$ and $\tau_{\text{intra}}$ denote the thresholds for inter-frame filtering and intra-frame merging, respectively. The budget allocation weights are denoted by $\lambda_u$ (uniform), $\lambda_s$ (static/structure), and $\lambda_e$ (dynamic/entropy).}
\label{tab:hyperparams}
\resizebox{\textwidth}{!}{%
\begin{tabular}{llcccl} 
\toprule
\textbf{Model} & \textbf{Benchmarks} & $\tau_{\text{inter}}$ & $\tau_{\text{intra}}$ & \textbf{Weights} ($\lambda_u, \lambda_s, \lambda_e$) & \textbf{Configuration Notes} \\
\midrule
\multirow{2}{*}{Qwen2-VL-7B} & VideoMME, MLVU & 0.20 & 0.55 & \multirow{2}{*}{0.8, 0.166, 0.034} & Balanced setting for standard duration. \\
 & EgoSchema & 0.30 & 0.75 & & Higher thresholds adopted for shorter duration. \\
\midrule
\multirow{2}{*}{Qwen2.5-VL-3B} & VideoMME, MLVU & 0.30 & 0.75 & \multirow{2}{*}{0.8, 0.166, 0.034} & Standard configuration. \\
 & EgoSchema & 0.50 & 0.75 & & Aggressive inter-frame filtering for redundancy. \\
\midrule
\multirow{6}{*}{LLaVA-OV-7B} & RVS, EgoSchema & 0.75 & 0.80 & 0.4, 0.575, 0.025 & YaRN scaling enabled (factor=5.5). \\
\cmidrule{2-6}
 & \multirow{5}{*}{VideoMME, MLVU} & \multicolumn{2}{c}{\multirow{2}{*}{\textit{Dynamic Adjustment}}} & \multirow{5}{*}{0.4, 0.575, 0.025} & \textbf{Duration-Adaptive Thresholds ($T$):} \\
 & & \multicolumn{2}{c}{} & & $\bullet$ $T \le 250\text{s}: \tau_{\text{inter}}=0.95, \tau_{\text{intra}}=0.97$ \\
 & & \multicolumn{2}{c}{\textit{based on Duration}} & & $\bullet$ $250\text{s} < T \le 1800\text{s}: \tau_{\text{inter}}=0.80, \tau_{\text{intra}}=0.85$ \\
 & & \multicolumn{2}{c}{} & & $\bullet$ $T > 1800\text{s}: \tau_{\text{inter}}=0.72, \tau_{\text{intra}}=0.82$ \\
 & & \multicolumn{2}{c}{} & & \small\textit{YaRN: factor=4.0 (VideoMME), 2.0 (MLVU).} \\
\midrule
\multirow{3}{*}{InternVL-3.5-2B} & MLVU & 0.50 & 0.75 & \multirow{3}{*}{0.3, 0.6, 0.1} & Input sampled to 64 frames. \\
 & VideoMME & 0.50 & 0.75 & & Input sampled to 100 frames. \\
 & EgoSchema & 0.75 & 0.75 & & Input sampled to 100 frames. \\
\bottomrule
\end{tabular}%
}
\end{table*}

\section{Detailed Experimental Setup and Additional Evaluations}
\label{app:ovo}

\subsection{Evaluation Protocols on RVS-Benchmarks}
For the streaming video understanding assessment on RVS-Ego and RVS-Movie \cite{di2025rekv}, we adhered to the standard evaluation protocols established in prior works \cite{zhang2024flash,di2025rekv,kim2025infinipot,yang2025streammem}. Specifically, we employed an LLM-based evaluator to judge the correctness of open-ended answers. 

\noindent \textit{Implementation Note regarding GPT Versions:} To maintain comparability with previous baselines, we targeted the standard `gpt-3.5-turbo' endpoint for evaluation. However, due to API provider updates, the response headers during our experiments identified the actual underlying backend model as `gpt-4o-mini-2025-07-18'. This detail is reported to ensure strict reproducibility of our results.

\subsection{Extended Evaluation on OVO-Bench}

To further evaluate the model's temporal reasoning capabilities—specifically Back-tracing, Real-time Visual Perception, and Proactive Future Response—we
assessed on OVO-Bench \cite{niu2025ovo}.

\begin{table*}[t]
\centering
\caption{Evaluation results on OVO-Bench. The table compares offline models and online streaming models separately. \textbf{Bold} indicates the best performance and \underline{underline} indicates the second best within each group.}
\definecolor{graybg}{rgb}{0.95, 0.95, 0.95}
\setlength{\tabcolsep}{2.8pt} 
\small 
\resizebox{\linewidth}{!}{
\begin{tabular}{l c ccccccc c ccc c ccc c c}
\toprule
\multirow{2}{*}{\textbf{Model}} & \multirow{2}{*}{\textbf{Frames}} &  AVG KV Size&\multicolumn{7}{c}{\textbf{Real-Time Visual Perception}} & \multicolumn{4}{c}{\textbf{Backward Tracing}} & \multicolumn{4}{c}{\textbf{Forward Active Responding}} & \multirow{2}{*}{\textbf{Overall}} \\
\cmidrule(lr){3-9} \cmidrule(lr){10-13} \cmidrule(lr){14-17}
 & &  &OCR & ACR & ATR & STU & FPD & OJR & Avg. & EPM & ASI & HLD & Avg. & REC & SSR & CRR & Avg. & Avg. \\

\midrule
 &\multicolumn{18}{c}{\textit{Methods w/o Compression}} \\
\midrule
LLaVA-NeXT-7B \cite{zhang2024llavanextvideo} & 64 &  $\approx$10K&\underline{69.8} & \textbf{59.6} & 66.4 & \textbf{50.6} & \textbf{72.3} & 61.4 & \underline{63.3} & 51.2 & 64.2 & 9.7 & 41.7 & 34.1 & \underline{67.6} & \textbf{60.8} & \textbf{54.2} & \textbf{53.1} \\
LLaVA-OneVision-7B \cite{li2024llava} & 64 &  $\approx$10K &67.1 & \underline{58.7} & \textbf{69.8} & 49.4 & \underline{71.3} & 60.3 & 62.8 & \textbf{52.5} & 58.8 & 23.7 & 45.0 & 24.8 & 66.9 & \textbf{60.8} & \underline{50.9} & \underline{52.9} \\
Qwen2-VL-7B \cite{wang2024qwen2} & 64 & $\approx$10K &69.1 & 53.2 & 63.8 & \textbf{50.6} & 66.3 & 60.9 & 60.7 & 44.4 & \textbf{66.9} & \textbf{34.4} & \textbf{48.6} & 30.1 & 65.7 & 50.8 & 48.9 & 52.7 \\
InternVL-V2-8B \cite{chen2024internvl} & 64 & $\approx$10K &68.5 & \underline{58.7} & \underline{69.0} & 44.9 & 67.3 & 56.0 & 60.7 & 43.1 & 61.5 & \underline{27.4} & 44.0 & 25.8 & 57.6 & 52.9 & 45.4 & 50.1 \\

Qwen2.5-VL-7B \cite{bai2025qwen2} & 1fps&  $\approx$50K &\textbf{74.5} & 57.8 & \textbf{69.8} & 50.0 & 70.3 & 58.2 & \textbf{63.4} & 49.8 & 65.5 & 26.9 & \underline{47.4} & \textbf{35.2} & 41.9 & 48.8 & 41.9 & 50.9 \\

\midrule
 &\multicolumn{18}{c}{\textit{Methods w/ Compression}} \\
\midrule

TimeChat-Online-7B \cite{yao2025timechat} & 1fps &  $\approx$ 5K&69.8 & 48.6 & \textbf{64.7} & 44.9 & \textbf{68.3} & 55.4 & 58.6 & \underline{53.9} & \textbf{62.8} & 9.1 & \underline{42.0} & 32.5 & 36.5 & 40.0 & 36.4 & 45.6 \\

\rowcolor{graybg}
\textbf{Qwen2.5-VL-7B + \textbf{Video-MemComp} (Ours)} & 1fps &  4K &\textbf{72.5} & \textbf{55.1} & 62.9 & \underline{52.8} & 66.3 & \textbf{60.3} & \textbf{61.7} & 48.5 & \underline{62.2} & \textbf{36.6} & \textbf{49.1} & \textbf{36.5} & 42.1 & 46.3 & 41.6 & \textbf{50.8} \\
\bottomrule
\end{tabular}
}

\label{tab:ovo_bench}
\end{table*}

We deployed our method on the Qwen2.5-VL-7B architecture. As shown in the table\ref{tab:ovo_bench}, our method achieved an overall score of 50.8, which is nearly on par with the full-input baseline Qwen2.5-VL-7B (50.9), while restricting the KV Cache to under 4000 tokens (hyperparameters were tuned for OVO-Bench on Qwen2.5-VL-7B). Of particular note, in the \textbf{Backward Tracing} task, our method achieved an average score of 49.1, surpassing the full baseline model's 47.4. This counter-intuitive result further proves that our memory retention mechanism does not simply discard information, but effectively distills and locks onto key cues from the historical stream, thereby achieving better-than-full-input performance in modeling long-term dependencies.

\subsection{Budget Allocation Hyperparameters Ablation Study Details}
\label{app:buget_ablation}
Table \ref{tab:full_ablation_all} presents the complete results of the hyperparameter ablation study on OVO-Bench. We categorize the 12 sub-tasks into three major groups: Backward Tracing, Real-time Perception, and Forward Active Responding.

\begin{table*}[t] 
\centering
\caption{\textbf{Comprehensive Ablation Study on Budget Allocation Hyperparameters.} We report the accuracy (\%) for all sub-tasks on OVO-Bench. The configurations are sorted by the Uniform Prior weight ($\lambda_u$). The extreme cases (pure strategies) are located at the top and bottom. The new global optimal configuration is highlighted in \textbf{bold} with a gray background.}
\label{tab:full_ablation_all}
\renewcommand{\arraystretch}{1.05} 
\setlength{\tabcolsep}{3pt} 
\resizebox{\textwidth}{!}{%
\begin{tabular}{ccc|cccc|ccccccc|cccc|c}
\toprule
\multicolumn{3}{c|}{\textbf{Hyperparams}} & \multicolumn{4}{c|}{\textbf{Backward Tracing}} & \multicolumn{7}{c|}{\textbf{Real-time Perception}} & \multicolumn{4}{c|}{\textbf{Forward Active}} & \textbf{Total} \\
$\lambda_u$ & $\lambda_s$ & $\lambda_e$ & EPM & HLD & ASI & \textit{Avg} & STU & OJR & ATR & FPD & ACR & OCR & \textit{Avg} & REC & SSR & CRR & \textit{Avg} & \textbf{Avg} \\
\midrule
0.00 & 0.00 & 1.00 & 48.82 & 34.95 & 62.16 & 48.64 & 49.44 & 55.98 & 62.93 & 66.34 & 55.96 & 69.13 & 59.96 & 37.48 & 42.44 & 42.08 & 40.67 & 49.76 \\
0.00 & 1.00 & 0.00 & 49.16 & 31.72 & 62.16 & 47.68 & 45.51 & 58.15 & 63.79 & 66.34 & 54.13 & 71.81 & 59.95 & 36.31 & 41.85 & 40.83 & 39.66 & 49.10 \\
\midrule
0.05 & 0.90 & 0.05 & 48.82 & 31.72 & 62.84 & 47.79 & 47.75 & 57.61 & 59.48 & 63.37 & 55.96 & 71.14 & 59.22 & 36.89 & 42.03 & 39.17 & 39.36 & 48.79 \\
0.10 & 0.90 & 0.00 & 48.82 & 31.72 & 62.16 & 47.57 & 47.19 & 61.41 & 65.52 & 69.31 & 55.05 & 71.81 & 61.71 & 36.21 & 36.99 & 43.75 & 38.99 & 49.42 \\
0.10 & 0.85 & 0.05 & 48.82 & 30.11 & 63.51 & 47.48 & 48.88 & 59.24 & 61.21 & 71.29 & 51.38 & 72.48 & 60.74 & 36.35 & 37.27 & 44.58 & 39.40 & 49.21 \\
0.10 & 0.80 & 0.10 & 48.82 & 31.72 & 59.46 & 46.67 & 45.51 & 63.04 & 61.21 & 67.33 & 52.29 & 73.83 & 60.53 & 36.16 & 37.84 & 42.92 & 38.97 & 48.72 \\
0.10 & 0.40 & 0.50 & 47.81 & 33.33 & 64.86 & 48.67 & 48.88 & 59.24 & 66.38 & 65.35 & 51.38 & 70.47 & 60.28 & 36.75 & 42.09 & 43.33 & 40.72 & 49.89 \\
\midrule
0.20 & 0.40 & 0.40 & 49.49 & 32.80 & 61.49 & 47.93 & 49.44 & 57.07 & 67.24 & 64.36 & 56.88 & 71.14 & 61.02 & 36.67 & 41.41 & 43.33 & 40.47 & 49.81 \\
0.245 & 0.625 & 0.13 & 47.47 & 31.72 & 63.51 & 47.57 & 48.31 & 57.07 & 65.52 & 65.35 & 55.96 & 71.14 & 60.56 & 36.91 & 41.45 & 42.08 & 40.15 & 49.42 \\
0.275 & 0.675 & 0.05 & 48.82 & 34.95 & 60.81 & 48.19 & 51.12 & 59.24 & 63.79 & 66.34 & 55.05 & 73.83 & 61.56 & 36.51 & 42.46 & 45.42 & 41.46 & 50.41 \\
\midrule
0.30 & 0.70 & 0.00 & 50.84 & 33.33 & 58.11 & 47.43 & 47.19 & 63.59 & 67.24 & 70.30 & 53.21 & 71.81 & 62.22 & 36.98 & 37.28 & 42.08 & 38.78 & 49.48 \\
0.30 & 0.675 & 0.025 & 48.48 & 36.56 & 62.16 & 49.07 & 52.81 & 59.24 & 62.93 & 66.34 & 55.05 & 71.81 & 61.36 & 36.09 & 42.01 & 46.25 & 41.45 & 50.63 \\
0.30 & 0.65 & 0.05 & 50.17 & 32.26 & 61.49 & 47.97 & 48.31 & 60.87 & 62.07 & 66.34 & 54.13 & 73.15 & 60.81 & 36.70 & 41.55 & 45.42 & 41.22 & 50.00 \\
0.30 & 0.60 & 0.10 & 51.18 & 31.72 & 60.14 & 47.68 & 45.51 & 61.41 & 65.52 & 69.31 & 52.29 & 74.50 & 61.42 & 36.51 & 38.80 & 40.42 & 38.58 & 49.23 \\
0.30 & 0.40 & 0.30 & 49.16 & 32.26 & 65.54 & 48.99 & 48.31 & 59.24 & 62.07 & 64.36 & 51.38 & 72.48 & 59.64 & 35.67 & 40.54 & 42.50 & 39.57 & 49.40 \\
0.30 & 0.30 & 0.40 & 49.16 & 33.33 & 62.16 & 48.22 & 48.31 & 58.70 & 62.07 & 65.35 & 55.96 & 67.79 & 59.70 & 36.55 & 41.74 & 43.75 & 40.68 & 49.53 \\
\rowcolor{gray!15} \textbf{0.35} & \textbf{0.625} & \textbf{0.025} & 51.52 & 35.48 & 62.16 & \textbf{49.72} & 51.69 & 61.96 & 62.07 & 67.33 & 55.05 & 71.14 & \textbf{61.54} & 35.96 & 42.52 & 45.00 & \textbf{41.16} & \textbf{50.81} \\
\midrule
0.50 & 0.50 & 0.00 & 51.52 & 31.72 & 62.84 & 48.69 & 48.31 & 59.78 & 61.21 & 73.27 & 51.38 & 73.15 & 61.18 & 36.47 & 36.56 & 40.42 & 37.82 & 49.23 \\
0.50 & 0.45 & 0.05 & 45.12 & 33.33 & 59.46 & 45.97 & 48.88 & 56.52 & 62.93 & 70.30 & 51.38 & 76.51 & 61.09 & 41.00 & 39.25 & 42.50 & 40.92 & 49.32 \\
0.50 & 0.40 & 0.10 & 49.49 & 34.41 & 62.16 & 48.69 & 51.12 & 56.52 & 63.79 & 66.34 & 55.05 & 68.46 & 60.21 & 36.27 & 41.77 & 44.58 & 40.88 & 49.93 \\
0.60 & 0.10 & 0.30 & 49.49 & 34.95 & 62.84 & 49.09 & 51.12 & 57.61 & 63.79 & 66.34 & 54.13 & 70.47 & 60.58 & 37.11 & 41.09 & 41.67 & 39.96 & 49.88 \\
0.60 & 0.05 & 0.35 & 48.48 & 34.41 & 62.84 & 48.58 & 48.31 & 55.98 & 66.38 & 66.34 & 54.13 & 71.81 & 60.49 & 35.86 & 40.95 & 43.33 & 40.05 & 49.71 \\
0.675 & 0.275 & 0.05 & 50.17 & 34.41 & 61.49 & 48.69 & 50.00 & 57.61 & 64.66 & 67.33 & 55.05 & 67.11 & 60.29 & 36.56 & 41.20 & 43.75 & 40.50 & 49.83 \\
\midrule
0.70 & 0.30 & 0.00 & 50.17 & 30.11 & 62.16 & 47.48 & 48.31 & 59.24 & 62.07 & 70.30 & 51.38 & 71.81 & 60.52 & 36.73 & 37.10 & 41.67 & 38.50 & 48.83 \\
0.70 & 0.25 & 0.05 & 50.17 & 34.41 & 60.81 & 48.46 & 50.00 & 57.07 & 65.52 & 66.34 & 55.96 & 67.57 & 60.41 & 36.35 & 41.16 & 42.92 & 40.14 & 49.67 \\
0.70 & 0.20 & 0.10 & 49.83 & 34.41 & 62.84 & 49.03 & 50.56 & 57.07 & 63.79 & 66.34 & 55.05 & 69.13 & 60.32 & 36.56 & 41.54 & 45.42 & 41.17 & 50.17 \\
1.00 & 0.00 & 0.00 & 48.82 & 34.95 & 62.84 & 48.87 & 50.00 & 55.43 & 62.07 & 66.34 & 55.05 & 69.13 & 59.67 & 36.45 & 41.68 & 42.50 & 40.21 & 49.58 \\
\bottomrule
\end{tabular}%
}
\end{table*}

\subsection{Token Compression Thresholds Ablation Study Details}
\label{app:sim_ablation}

Table \ref{tab:sim_ablation_data} lists detailed performance metrics for all evaluated combinations of inter-frame and intra-frame similarity thresholds.
\begin{table*}[!t]
\centering
\caption{\textbf{Ablation Study of Token Compression Thresholds on OVO-Bench.} We report accuracy (\%) for Backward, Real-time, and Forward tasks. The best configuration is highlighted in \textbf{bold} and gray background.}
\label{tab:sim_ablation_data}
\resizebox{\textwidth}{!}{%
\begin{tabular}{cc|cccc|ccccccc|cccc|c}
\toprule
\multicolumn{2}{c|}{\textbf{Thresholds}} & \multicolumn{4}{c|}{\textbf{Backward Tracing}} & \multicolumn{7}{c|}{\textbf{Real-time Perception}} & \multicolumn{4}{c|}{\textbf{Forward Active}} & \textbf{Total} \\
$\tau_{\text{inter}}$ & $\tau_{\text{intra}}$ & EPM & HLD & ASI & \textit{Avg} & STU & OJR & ATR & FPD & ACR & OCR & \textit{Avg} & REC & SSR & CRR & \textit{Avg} & \textbf{Avg} \\
\midrule
0.25 & 0.75 & 43.77 & 31.72 & 58.78 & 44.76 & 43.26 & 54.89 & 62.07 & 68.32 & 55.96 & 65.10 & 58.27 & 31.09 & 41.65 & 45.42 & 39.38 & 47.47 \\
0.30 & 0.60 & 45.12 & 29.57 & 57.43 & 44.04 & 42.13 & 53.26 & 62.93 & 68.32 & 51.38 & 65.77 & 57.30 & 29.39 & 41.49 & 54.58 & 41.82 & 47.72 \\
0.30 & 0.75 & 46.46 & 29.57 & 58.11 & 44.71 & 43.82 & 54.89 & 62.07 & 67.33 & 55.05 & 67.11 & 58.38 & 36.73 & 40.97 & 44.17 & 40.62 & 47.91 \\
0.30 & 0.95 & 45.45 & 31.18 & 60.14 & 45.59 & 42.70 & 53.26 & 59.48 & 66.34 & 54.13 & 67.79 & 57.28 & 33.07 & 41.99 & 40.00 & 38.35 & 47.08 \\
0.50 & 0.75 & 46.13 & 29.03 & 60.81 & 45.32 & 46.63 & 54.89 & 58.62 & 70.30 & 57.80 & 63.09 & 58.55 & 33.34 & 42.74 & 47.66 & 41.25 & 48.37 \\
0.50 & 0.80 & 48.82 & 32.80 & 62.16 & 47.93 & 44.94 & 60.33 & 62.93 & 67.33 & 51.38 & 65.77 & 58.78 & 34.04 & 43.13 & 42.50 & 39.89 & 48.87 \\
0.50 & 0.95 & 48.15 & 30.65 & 61.49 & 46.76 & 43.82 & 57.07 & 62.07 & 66.34 & 55.05 & 69.80 & 59.02 & 35.23 & 41.39 & 42.50 & 39.71 & 48.50 \\
0.60 & 0.85 & 48.82 & 32.26 & 61.49 & 47.52 & 45.51 & 55.98 & 64.66 & 66.34 & 55.05 & 71.81 & 59.89 & 35.66 & 41.05 & 39.58 & 38.76 & 48.72 \\
0.70 & 0.80 & 49.16 & 33.87 & 62.16 & 48.40 & 48.88 & 54.35 & 68.10 & 66.34 & 55.96 & 68.46 & 60.35 & 34.80 & 41.48 & 45.42 & 40.57 & 49.77 \\
0.70 & 0.85 & 48.82 & 33.33 & 62.16 & 48.11 & 43.82 & 55.98 & 63.79 & 66.34 & 57.80 & 67.11 & 59.14 & 36.10 & 41.83 & 41.25 & 39.73 & 48.99 \\
0.75 & 0.80 & 49.16 & 34.95 & 61.49 & 48.53 & 50.00 & 56.52 & 62.07 & 66.34 & 54.13 & 69.13 & 59.70 & 35.65 & 42.82 & 42.92 & 40.46 & 49.56 \\
0.80 & 0.60 & 46.46 & 28.49 & 57.43 & 44.13 & 47.19 & 54.89 & 62.07 & 70.30 & 55.05 & 62.42 & 58.65 & 27.06 & 38.91 & 50.83 & 38.93 & 47.24 \\
0.80 & 0.75 & 48.48 & 30.65 & 60.14 & 46.42 & 47.75 & 55.43 & 63.79 & 66.34 & 53.21 & 67.11 & 58.94 & 34.52 & 40.49 & 48.75 & 41.25 & 48.87 \\
\rowcolor{gray!15} \textbf{0.80} & \textbf{0.85} & 48.48 & 36.56 & 62.16 & \textbf{49.07} & 52.81 & 60.33 & 62.93 & 66.34 & 55.05 & 72.48 & \textbf{61.66} & 36.50 & 42.05 & 46.25 & \textbf{41.60} & \textbf{50.77} \\
0.90 & 0.80 & 47.81 & 33.87 & 64.19 & 48.62 & 46.63 & 54.89 & 59.48 & 66.34 & 55.05 & 71.81 & 59.03 & 35.25 & 42.33 & 44.17 & 40.58 & 49.41 \\
0.95 & 0.90 & 48.15 & 31.72 & 64.19 & 48.02 & 48.31 & 57.07 & 62.93 & 66.34 & 55.96 & 68.46 & 59.84 & 36.43 & 38.95 & 42.50 & 39.29 & 49.05 \\
\bottomrule
\end{tabular}%
}
\end{table*}

\end{document}